\documentclass[final]{siamltex}

\usepackage{latexsym, graphicx, epsfig, amsmath, amsfonts,amssymb,bm}
\usepackage{caption, subcaption}
\usepackage{color,soul}
\usepackage{epstopdf}
\usepackage{caption}
\usepackage{enumitem}  
\usepackage{xcolor}
\usepackage{algorithm}
\usepackage[noend]{algpseudocode}

\algrenewcommand\algorithmicensure{\textbf{Output:}}

\newcommand{\rN}[1]{r^{(#1)}}
\newcommand{\mahn}[1]{\left\|#1\right\|^2_{\Sigma^{-1}}}

\usepackage{booktabs}
\usepackage{hyphenat}
\usepackage{multirow}
\usepackage{array}
\usepackage{color}
\usepackage{url}
\usepackage{comment}
\usepackage{diagbox}
\usepackage{graphicx}
\usepackage[label font=bf,labelformat=simple]{subfig}
\usepackage{caption}
\allowdisplaybreaks

\def\be{\begin{equation}}
\def\ee{\end{equation}}

\def\0{\mathbf{0}}

\title{Bifidelity Parameter Estimation Using Conditional Diffusion Models}
\title{Bifidelity Parameter Estimation Using Conditional Diffusion Models\thanks{This manuscript has been authored by UT-Battelle, LLC, under contract DE-AC05-00OR22725 with the US Department of Energy (DOE). The US government retains and the publisher, by accepting the article for publication, acknowledges that the US government retains a nonexclusive, paid-up, irrevocable, worldwide license to publish or reproduce the published form of this manuscript, or allow others to do so, for US government purposes. DOE will provide public access to these results of federally sponsored research in accordance with the DOE Public Access Plan. The work of C.Tatsuoka and D.Xiu was partially supported by AFOSR FA9550-22-1-0011.}}

\author{
  Caroline Tatsuoka\thanks{Department of Mathematics, The Ohio State University, Columbus, OH 43210, USA (tatsuoka.3@osu.edu, xiu.16@osu.edu).}
  \and 
  Minglei Yang\thanks{Fusion Energy Science Division, Oak Ridge National Laboratory, Oak Ridge, TN, 37831, USA (yangm@ornl.gov).}
  \and 
  Dongbin Xiu\footnotemark[1]  
  \and 
  Guannan Zhang\thanks{Computer Science and Mathematics Division, Oak Ridge National Laboratory, Oak Ridge, TN, 37831, USA (zhangg@ornl.gov).}
}

\begin{document}
\maketitle

\begin{abstract} We present a bifidelity method for uncertainty quantification of parameter estimates in complex systems, leveraging generative models trained to sample the target conditional distribution. In the Bayesian inference setting, traditional parameter estimation methods rely on repeated simulations of potentially expensive forward models to determine the posterior distribution of the parameter values, which may result in computationally intractable workflows. Furthermore, methods such as Markov Chain Monte Carlo (MCMC) necessitate rerunning the entire algorithm for each new data observation, further increasing the computational burden. Hence, we propose a novel method for efficiently obtaining posterior distributions of parameter estimates for high-fidelity models given data observations of interest. The method first constructs a low-fidelity, conditional generative model capable of amortized Bayesian inference and hence rapid posterior density approximation over a wide-range of data observations. When higher accuracy is needed for a specific data observation, the method employs adaptive refinement of the density approximation. It uses outputs from the low-fidelity generative model to refine the parameter sampling space, ensuring efficient use of the computationally expensive high-fidelity solver. Subsequently, a high-fidelity, unconditional generative model is trained to achieve greater accuracy in the target posterior distribution. Both low- and high- fidelity generative models enable efficient sampling from the target posterior and do not require repeated simulation of the high-fidelity forward model. We demonstrate the effectiveness of the proposed method on several numerical examples, including cases with multi-modal densities, as well as an application in plasma physics for a runaway electron simulation model.\end{abstract}


%





%
%
%
%


\section{Introduction}\label{sec:intro}

Uncertainty quantification (UQ) of parameter estimates in complex systems is critical for comprehensive and interpretable model calibration. In practice, these estimates are frequently derived from incomplete, noisy, or sparse data, either due to inherent limitations in the modeling or constraints in data collection. For complex systems—where underlying behaviors may be highly nonlinear, chaotic, or stochastic—it becomes essential to understand the uncertainty associated with the data-driven parameterizations. Bayesian inference, a powerful probabilistic framework, can be utilized in these potentially ill-posed settings for approximating the conditional distribution of the parameters given the available data.

Traditional methods for parameter estimation such as MCMC \cite{hastings1970monte}, require a tractable likelihood and prior density function; the forward physical model is embedded within the likelihood and hence thousands or potentially millions of forward model evaluations may be required in order to achieve convergence to the desired parameter posterior distribution. This method may become intractable when a forward evaluation of the underlying system is computationally expensive, requiring excessive run-times to converge. Alternative approaches such as variational inference \cite{blei2017variational} similarly require tractable likelihoods and repeated simulations during optimization of the evidence lower bound (ELBO) for obtaining the posterior density approximation. 
Other likelihood-free methods, such as approximate Bayesian computation (ABC), rely on rejection algorithms based on summary statistics of the forward model, again requiring a large number of forward simulation runs \cite{sisson2018handbook}. 

{To address these limitations, multifidelity modeling leverages both low- and high-fidelity models to reduce the computational burden while maintaining accuracy. In conjunction with MCMC, \cite{cui2015data} builds data-driven, reduced-order surrogate models for Bayesian inverse problems, while \cite{peherstorfer2019transport} proposes a multifidelity preconditioned MCMC method that uses low-fidelity models to build a transport-map based proposal distribution, enabling efficient sampling that maintains a stationary distribution corresponding to the high-fidelity model. Rather than sampling from the full posterior distribution, alternative multifidelity methods directly estimate specific quantities of interest (QoIs) through efficient estimators. Multifidelity importance sampling (MFIS) uses a surrogate to construct a biasing distribution and combines many cheap surrogate-based draws with a small number of high-fidelity evaluations to estimate QoIs; \cite{alsup2023context} extends MFIS to Bayesian inverse problems, learning context-aware surrogates that adaptively select the fidelity level to minimize cost while meeting user-specified accuracy tolerances for the MFIS estimator.}

{Multilevel methods in scientific computing leverage a hierarchy of forward model discretizations to balance accuracy and efficiency. In \cite{dodwell2019multilevel}, the authors develop a multilevel Metropolis-Hastings algorithm using a hierarchy of computational models, significantly reducing computational cost. \cite{alsup2022multilevel} extends Stein Variational Gradient Descent to a multilevel method for Bayesian inverse problems that leverages distributions from different discretization levels of the forward model, using cheaper coarse levels to precondition particle transport and reduce the overall computational cost of approximating the target distribution.} {\cite{latz2018multilevel} approximates posterior distributions in Bayesian inverse problems via a sequential Monte Carlo (SMC) sampling that exploits a hierarchy of PDE discretizations, adaptively alternating between tempering (to gradually introduce the likelihood) and bridging (to transition across discretization levels) for computationally efficient sampling of the parameters' posterior. In \cite{farcas2020multilevel}, a sampling-free multilevel Bayesian inversion method is introduced, where posterior-focused surrogates are built adaptively with sparse grids using weighted Leja points, reducing the number of expensive forward model evaluations.} 

Recently in the machine learning community, there is growing interest in using machine learning methods for  uncertainty-quantification tasks, which can help mitigate the described efficiency issues while providing a robust uncertainty quantification on parameter estimates. One approach is to introduce fast surrogate models, such as trained neural networks, to accelerate forward model evaluations within MCMC \cite{asher2015review, dunbar2021calibration, weber2020deep, xi2017calibration}. While potentially reducing computationally costs, the surrogate model must be globally accurate across the entire parameter domain to ensure reliable posterior samples, and hence sufficient training of such a surrogate model can become a challenge when the available data is sparse, noisy or incomplete. Generative modeling is a form of ML which has seen success in many applications, including image processing \cite{dhariwal2021diffusion, goodfellow2014generative, ho2020denoising, song2019generative, song2020score}, anomaly detection \cite{mimikos2024score,schlegl2017unsupervised} and natural language processing \cite{austin2021structured, hoogeboom2021argmax, ma2019flowseq}. Generative modeling architectures include variational auto-encoders (VAEs), generative adverserial networks (GANs), normalizing flows, and the diffusion model. VAEs enable generative modeling by using encoder-decoder architectures to transform samples to their low-dimensional, latent space representations for learning of the latent probability distribution, which is then used for sampling in parameter space via the decoder \cite{cheng2024bi,kingma2013auto,xu2024modeling}. This approach, however, may lead to a loss of interpretability and potential mode collapse due to insufficient information retained in the latent space. Normalizing flows have been extensively applied in uncertainty quantification (UQ) \cite{khorashadizadeh2023conditional,lu2022invertible,papamakarios2021normalizing,wang2023efficient,yang2024pseudoreversible}, leveraging invertible and differentiable transformations for mapping a standard Gaussian distribution to a desired target distribution. Despite the many advancements in flow designs, the requirement of tractable inverse maps and determinant of the maps' Jacobian matrix may limit the expressivity of these models and also result in high computational costs. GANs have also seen success for posterior sample generation \cite{chen2024learning, goodfellow2014generative, patel2021gan}, however, the unsupervised training for optimization of the adversarial loss function can lead to mode collapse and may require entropic regularization \cite{baptista2024conditional,li2018learning,salimans2016improved,yang2019adversarial}. 

Diffusion models learn to generate samples from the target distribution by sequentially injecting noise into the available data samples, and subsequently learning to de-noise the corrupted data in a reverse process. It has seen a wide-range of extensions and has been successful in applications, see \cite{baldassari2024taming, li2023diffusion,qiao2024score,sun2024provable}. Particularly, score-based diffusion models, presented in \cite{song2020score}, has been adapted to UQ for complex systems \cite{bao2023scorebased,liu2024diffusion, lu2024diffusion}, using an efficient Monte Carlo estimator of the score function, allowing for a training-free diffusion model. While score-based diffusion models traditionally use a neural network to approximate the score function, the lack of labeled data mapping samples from the initial distribution to samples from the terminal distribution within the stochastic process motivates the use of unsupervised learning algorithms, in turn requiring the storage of numerous stochastic paths of the forward SDE and increasing computational costs. Instead, by introducing a Monte Carlo estimator, one can efficiently learn the score-function at each temporal location when solving the reverse-time process. The reverse-time SDE is additionally replaced by the corresponding reverse-time ODE \cite{maoutsa2020interacting}, allowing for a deterministic mapping between initial and terminal states of the samples. This enables the supervised learning of a generative model, taken to be a simple feed forward neural network. After training, the generative model can act as an efficient approximation tool of the target posterior density \cite{lu2024diffusion}. This method was further extended in  \cite{fan2024genai4uq,lu2024diffusion} for amortized Bayesian inference \cite{zammit2024neural}, allowing for posterior density approximations over a wide-range of observations. 


Informative sampling of the parameter space 
becomes imperative for efficient use of computational resources when attempting to accurately approximate the posterior distribution of the high-fidelity model's parameter values. Hence, we propose the following bifidelity framework: first, a low-fidelity generative model is developed to perform amortized Bayesian inference, offering fast and efficient approximation of target posterior distributions across a wide range of observation values through a single neural network. The low-fidelity generative model's ability to conduct rapid posterior density approximations resolves much of the computational burden associated with many Bayesian methodologies {such as MCMC or sequential Monte Carlo sampling and their multifidelity and multilevel extensions which require restarting the algorithm for every new observation of interest}.  
The initial prior dataset is considered low-fidelity due to the sparsity of data samples and/or due to the low-resolution numerical solvers used to generate data {In the latter setting, our method can be viewed as a two-level Bayesian inference method}. If the results of the low-fidelity approximation are sufficient for a specific data observation of interest, the user does not need to build the high-fidelity model. However, if a higher-fidelity approximation is required, the low-fidelity model enables refined sampling of the parameter space. This reduces the overall computational overhead by narrowing the sampling domain to regions of high probability of the target posterior density, {as in the aforementioned adaptive multifidelity and multilevel algorithms which aim to utilize models of varying fidelity to concentrate computational effort in regions of high posterior probability}. The high-fidelity solver then uses the refined data samples to generate output values and train the high-fidelity, unconditional generative model. The proposed two-step method combines efficient posterior sample generation with adaptive refinement of posterior density approximations, allowing for both speed and accuracy.


{Hence, the main contributions of this work are as follows: (i) we develop a bifidelity generative modeling framework that integrates amortized Bayesian inference with adaptive refinement of posterior density approximations, demonstrating how low-fidelity models can guide informative sampling, reducing the cost of high-fidelity Bayesian inference; and (ii) we validate the framework with several numerical examples, demonstrating that it achieves both computational efficiency and accuracy when compared to MCMC and SMC sampling approaches.}

The remainder of the paper is organized as follows: in Section \ref{sec:prob} we present the problem setting and motivation. In Section \ref{sec:method}, we review preliminaries, specifically recalling the training-free diffusion model methodologies proposed in \cite{liu2024diffusion, lu2024diffusion}, and present the proposed method. {Section \ref{sec:num} presents numerical examples to demonstrate the effectiveness of the method, and in Section \ref{sec:conclusion} we provide concluding remarks.} 

\section{Problem setting}\label{sec:prob}

We setup the parameter estimation problem for a general physical model. 
Let \(x \in \Omega \subset \mathbb{R}^{s}\) be the spatial variable with the support \(\Omega\) a bounded domain with a sufficiently smooth boundary, and let \(T>0\) be a finite time horizon. We are interested in estimating the parameter, denoted by $\theta \in \mathbb{R}^d$, of the following differential equation:
\begin{equation}\label{eq:system}
\mathcal{L}[u(x,t, \theta);\theta] = 0, \quad x \in \Omega,\ t \in [0,T],
\end{equation}
where \(\mathcal{L}\) is a (nonlinear) differential operator and $u(x,t,\theta)$ is the solution of the differential equation belonging to an appropriate Sobolev space, such as \(H^1(\Omega)\), ensuring sufficient regularity. The parameter \(\theta \in \Gamma \subset \mathbb{R}^{d}\) is a $d$-dimensional vector defined in a compact set \(\Gamma \subset \mathbb{R}^{d}\). We assume that for each \(\theta \in \Gamma\), there exists a unique solution \(u(x,t,\theta)\) that satisfies the differential equation \(\mathcal{L}[u(x,t,\theta);\theta]=0\), subject to suitable initial and boundary conditions. This uniqueness and well-posedness property is a standard assumption, often guaranteed by classical results in partial differential equations and dynamical systems theory.

In real-world applications, the parameter $\theta$ is not directly observable, we need to estimate $\theta$ using information obtained via an observation operator, denoted by
\begin{equation}\label{eq:obs_operator}
y = \mathcal{H}(u(x,t,\theta)) + \varepsilon
\end{equation}
 {where the mapping \(\mathcal{H}: H^1({\Omega}) \rightarrow \mathbb{R}^q\) associates the solution \(u(\cdot, \cdot, \theta)\) to an observable quantity and $\varepsilon \sim \mathcal{N}(0, \Sigma)$ is the Gaussian additive observation noise}. This operator often represents measurements of the system, such as sensor outputs, integral measurements over subdomains, or pointwise evaluations at specified spatial locations. We assume observation data is noisy, denoted by


The parameter estimation problem is defined in the Bayesian manner. Specifically, 
we assume that the parameter $\theta$ has a prior probability density function, denoted by $p_\Theta(\theta)$ supported on $\Gamma$, and a likelihood function corresponding to the observation operator in Eq.~\eqref{eq:obs_operator}, i.e.,
\begin{equation}\label{eq:likelihood}
p_{Y |\Theta}(y |\theta) \propto \exp\left[-\frac{1}{2}\big(y - \mathcal{H}(u(x,t,\theta))\big)^{\top} \Sigma^{-1}\big(y - \mathcal{H}(u(x,t,\theta))\big)\right],
\end{equation}
where $\Theta$ and $Y$ are the random variables corresponding to the model parameter $\theta$ and the observable quantity $y$, respectively, and 
\(\Sigma\) is the covariance matrix characterizing the noise statistics in the measurements. Given an observation \(Y = y\), the posterior distribution of the parameter $\theta$ is obtained through Bayes' theorem:
\begin{equation}\label{eq:bayes}
p_{\Theta|Y}(\theta |  y) \propto  p_{Y|\Theta}( y | \theta) \; p_{\Theta}(\theta),
\end{equation}
encapsulates both the observational information and the prior assumptions on the unknown parameters. In practice, this posterior distribution often forms the basis for various Bayesian inference techniques, including Markov Chain Monte Carlo sampling, variational inference, and surrogate modeling approaches.

Quantifying the resulting uncertainty in input or parameter information is essential for predictive modeling. However, this uncertainty quantification becomes computationally challenging when the numerical solver for solution $u$ in Eq.~\eqref{eq:system} is expensive. The Bayesian approach to inverse problems provides a rigorous foundation for inference from noisy and incomplete data, with Markov Chain Monte Carlo (MCMC) being particularly valuable for sampling from posterior distributions that often lack closed-form solutions. However, a significant limitation arises: for each new observation $y$, one must re-run the entire MCMC sampling procedure.
To overcome this computational challenge, we propose the development of a bifidelity generative model that enables amortized Bayesian inference through score-based diffusion models. For any observation $y\sim p_{Y}(y)$, our proposed generative model can directly sample from the posterior distribution $p_{\Theta|Y}$ without requiring repeated simulations. The framework offers both low- and high-fidelity sampling options, with the high-fidelity option requiring only an additional one-time training cost. The detailed architecture and implementation of this bifidelity generative model will be discussed in Section \ref{sec:method}.

\section{bifidelity parameter estimation using conditional diffusion models}\label{sec:method}
\begin{figure}[h!]
    \centering
    \includegraphics[width=0.9\linewidth]{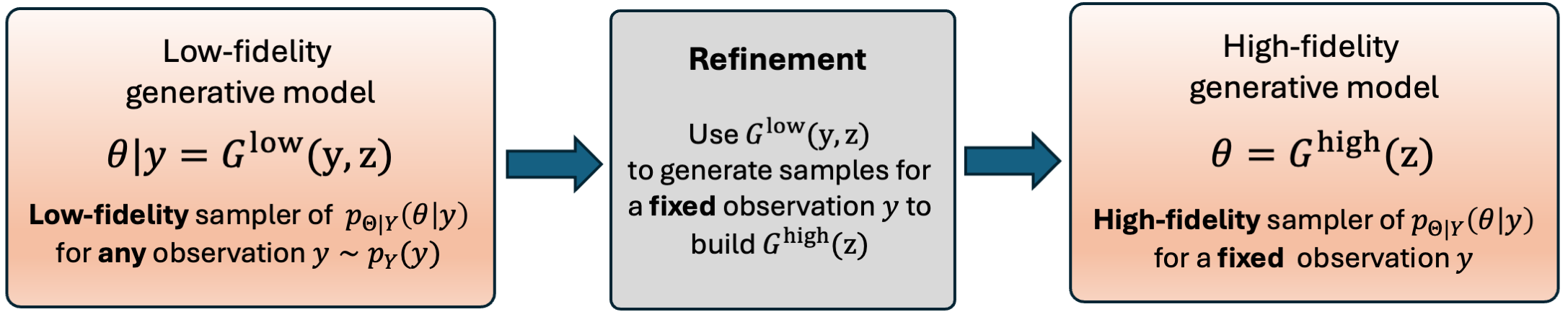}

    \caption{Illustration of the proposed bifidelity parameter estimation framework. The low-fidelity generative model efficiently approximates the posterior distribution $p_{\Theta|Y}(\theta | y)$ for any observation $y \sim p_Y(y)$, offering substantial computational advantages over traditional MCMC sampling which requires new runs for each observation. When $G^{\rm low}(y,z)$ cannot achieve the desired accuracy for a specific observation, it can still guide the sampling process by identifying high-probability regions of $p_{\Theta|Y}(\theta | y)$. These samples then serve as additional training data for constructing the more accurate high-fidelity generative model $G^{\rm high}(z)$.}
    \label{fig:workflow}
\end{figure}

In this paper, we propose a novel bifidelity framework for parameter estimation that addresses the key computational challenges in Bayesian inference. Our approach involves two generative models, denoted by
\begin{equation}\label{eq:low-high}
    \theta|y = G^{\rm low}(y, z) \;\; \text{ and }\;\; 
    \theta = G^{\rm high}(z), 
\end{equation}
where $G^{\rm low}$ is a low-fidelity generative model, $G^{\rm high}$ a high-fidelity generative model, and $z \in \mathbb{R}^d$ a random variable with the standard normal distribution $\mathcal{N}(0, \mathbf{I}_d)$. The low-fidelity generative model $G^{\rm low}(y,z)$ provides a computationally efficient approach to approximate the posterior distribution $p_{\Theta|Y}(\theta | y)$ across the entire space of possible observations $y \sim p_Y(y)$. Unlike traditional MCMC methods that require new sampling chains for each new observation, the low-fidelity model provides rapid posterior approximations through a single trained network. While $G^{\rm low}(y,z)$ may not achieve the desired accuracy threshold for certain observations, its ability to identify high-probability regions of $p_{\Theta|Y}(\theta | y)$ can be used as a good proposal distribution, 
effectively narrowing the search space for more accurate sampling of $p_{\Theta|Y}(\theta | y)$ for a fixed $y$. The samples generated from these high-probability regions can be used as additional training data for constructing the high-fidelity generative model $G^{\rm high}(z)$, ensuring that computational resources are focused on the most relevant regions of the parameter space. The proposed workflow is illustrated in Figure \ref{fig:workflow}. 
\vspace{0.2cm}

The key innovation of the proposed method lies in its adaptive nature. The low-fidelity model serves as an initial rapid approximation tool, capable of handling diverse observation scenarios without the computational overhead typically associated with MCMC methods. This initial approximation is particularly useful in scenarios requiring real-time parameter estimation or when dealing with large-scale inference problems. The high-fidelity model of our method activates selectively, targeting cases where enhanced accuracy is crucial. By utilizing the low-fidelity model's output to refine the sampling space, we can reduce the computational burden typically associated with the high-fidelity modeling. 
In this work, we use the training-free score-based diffusion model \cite{liu2024diffusion,liu2024trainingfreecondition,bao2024ensemble} to build both the low-fidelity and high-fidelity generative models. Section \ref{sec:overview_diff} provides a brief overview of the score-based diffusion model. Section \ref{sec:low} introduces the construction of a low-fidelity generative model using a conditional diffusion model with a training-free score estimator. Section \ref{sec:high} describes how to exploit the low-fidelity generative model as a proposal distribution to build a high-fidelity generative model.

%


\subsection{Overview of the score-based diffusion model}\label{sec:overview_diff} 
The score-based diffusion model consists of a forward process, which transforms samples from a target distribution to pure noise, and reverse process, which transforms the noise back to samples from the target distribution. Let $Z_\tau \in \mathbb{R}^d$ denote the forward process defined over a bounded pseudo-temporal domain $\tau \in [0,1]$, where the initial distribution $Z_0$ is the posterior distribution $p_{\Theta|Y}(\theta|y)$ from Eq.~\eqref{eq:bayes}. The forward process is given by the following linear SDE,
\begin{equation} \label{forward_SDE}
    dZ_{\tau} = b({\tau})Z_{\tau} d\tau + \sigma(\tau)dW_{\tau}, \quad \text{with} \quad Z_{0} = \Theta|Y  \;\; \text{and} \;\; Z_1 = Z,
\end{equation}
where $W_\tau$ is the standard Brownian motion. By choosing the appropriate drift and diffusion coefficients $b(\tau)$ and $\sigma_\tau$ as
\begin{equation} \label{drift_diffusion}
    b(\tau) = \frac{d \log \alpha_{\tau}}{d \tau}, \quad \sigma^2(\tau) = \frac{d \beta^2_{\tau}}{d\tau} - 2 \frac{d \log \alpha_\tau}{d \tau} \beta^2_{\tau}\quad 
\text{with}\;\; \alpha_{\tau} = 1 - \tau, \; \beta^2_{\tau} = \tau,
\end{equation}
it follows that the forward diffusion kernel for a fixed $Z_0 = z_0$ is defined as the following Gaussian distribution,
\begin{equation}\label{eq:forward_diffusion_kernel}
    p_{Z_t|Z_0}(z_t|z_0) = \phi(\alpha_\tau z_0, \beta^2_\tau\mathbf{I}_d) \;\text{for}\; \tau \in [0,1],
\end{equation}
where here $\phi$ denotes Gaussian density function with mean $\alpha_\tau$ and covariance matrix $\beta^2_\tau\mathbf{I}_d$. The forward SDE transports the target conditional distribution $Z_0 | Y$ to the standard normal $Z_1 \sim \mathcal{N}(0, \mathbf{I}_d)$ \cite{song2020score,bao2024ensemble}.

The score function  $S(z_{\tau},\tau)$ of the forward SDE is defined as
\begin{equation} \label{score_function}
    S(z_\tau, \tau):= \nabla_{z_\tau} \log q_{Z_\tau}(z_{\tau})
\end{equation}
where $q_{Z_\tau}$ is the probability density function of the state $Z_\tau$ at time $\tau$ in the forward SDE in Eq.~\eqref{forward_SDE}.

{It is shown in \cite{maoutsa2020interacting} the reverse process, which shares the same time-marginal density as the forward SDE in Eq.~\eqref{forward_SDE}, can be obtained by the following probability flow ordinary differential equation (ODE)}
%
%


\begin{equation} \label{eq:reverse_ODE}
    dZ_{\tau} = \left[b(\tau)Z_{\tau} - \frac{1}{2}\sigma^2(\tau)S(Z_{\tau},\tau) \right] dt \quad \text{with} \quad Z_0 = \Theta|Y \;\; \text{and} \;\; Z_1 = Z.
\end{equation}

This reverse ODE allows for the generation of new samples from the target distribution by transporting samples from the standard normal distribution at $\tau = 1$ to the target distribution at $\tau = 0$.  

The key step is the approximation of the score function; while traditionally approximated by a neural network, this can be a computationally intensive task as it relies on an unsupervised learning algorithm, requiring many iterations of forward SDE paths for learning of the score function. Thus, we utilize the training-free Monte Carlo estimator approach for approximating the score function, as described in step 2 of the following section.

\subsection{Construction of the low-fidelity generative model}\label{sec:low}
This section describes how to use the score-based diffusion model to construct the low-fidelity generative model in Eq.~\eqref{eq:low-high}. The procedure consists of three steps. 

\paragraph{\bf\em Step 1: Collecting samples of the joint distribution $p_{\Theta, Y}(\theta, y)$} 
In this work, we assume that all samples are generated by solving the underlying differential equation in Eq.~\eqref{eq:system}. We first generate a set of samples, denoted by $\{\theta^{(n)}\}_{n=1}^{N_{\rm prior}}$ from the prior distribution $p_\Theta (\theta)$ in Eq.~\eqref{eq:bayes}. {For each $\theta^{(n)}$, we evaluate Eq.~\eqref{eq:obs_operator} with $\mathcal{H}(u(x,t,\theta^{(n)}))$ to obtain the corresponding observation value $y^{(n)}$.} The pair $(\theta^{(n)},y^{(n)})$ is one sample from the joint distribution $p_{\Theta, Y}(\theta, y)$, and the set of prior samples is defined by
\begin{equation}\label{eq:prior_sample} 
\mathcal{S}_{\rm prior} = \left\{\left(\theta^{(n)},y^{(n)}\right)\right\}_{n=1}^{N_{\rm prior}}. 
\end{equation} 
{The prior sample set $\mathcal{S}_{\rm prior}$ is regarded as a low-fidelity sample set under one of two possible scenarios. First, to reduce the cost of solving the underlying differential equation, we could use a low-fidelity numerical solver (e.g., with low-order spatial/temporal discretization schemes) to generate the observation data $y^{(n)}$, enabling us to afford more prior data for denser coverage of the prior domain $\Gamma$. }


 The sparsity or overly large distance between samples in the prior data set can lead to over-estimation (low-fidelity estimation) of the uncertainty of the posterior distribution.

\paragraph{\bf\em Step 2: Generating labeled data using the training-free diffusion model}

This step uses the prior data set $\mathcal{S}_{\rm prior}$ in Eq.~\eqref{eq:prior_sample} to generate labeled data for supervised learning of the low-fidelity model $G^{\rm low}(y, z)$ \eqref{eq:low-high}, to be used in the next step. 


Our goal is to build the following labeled data set,
\begin{equation}\label{eq:label_data}
   \mathcal{S}_{\rm label} = \left\{\left(y^{(m)}, z^{(m)}, \theta^{(m)}\right): m = 1,\dots,M\right\},
\end{equation}
where $y^{(m)}$ is a sample from $p_Y(y)$, $z^{(m)}$ is a sample from $\mathcal{N}(0, \mathbf{I}_d)$, and $\theta^{(m)}$ is a sample from the conditional distribution $p_{\Theta|Y}(\theta|y)$.
Compared to Eq.~\eqref{eq:low-high}, we can see that $(y^{(m)},z^{(m)})$ is the input and $\theta^{(m)}$ is the output of the low-fidelity model. 

{As discussed in \cite{liu2024diffusion,lu2024diffusion}, employing the probability flow ODE in Eq.~\eqref{eq:reverse_ODE} yields a deterministic mapping from the standard normal distribution samples at $\tau=1$ to posterior samples at $\tau=0$. Within $\mathcal{S}_{\rm label}$, each posterior sample is paired with its corresponding Gaussian noise sample, providing supervised training data for the efficient sampler $G^{\rm low}(y,z)$. The conditioning variable $y$ serves as an additional label in this mapping, enabling supervised learning of the conditional generator.}

{We break this step into two parts: in step 2 (a), we first describe the training-free Monte Carlo approximation of the score function, and in step 2 (b), we detail the construction of $\mathcal{S}_{\rm label}$. }


\paragraph{\bf\em {Step 2 (a): Score function approximation}}
{The reverse ODE defined in Eq.~\eqref{eq:reverse_ODE} requires an approximation of the score function defined in Eq.~\eqref{score_function}.} 
To obtain this approximation, we first re-write the probability density function of $q_{Z_\tau}(z_{\tau})$ in Eq.~\eqref{score_function} as an integral, i.e.,
\begin{equation}\label{forward_states_pdf}
    q_{Z_\tau}(z_\tau) = \int_{\mathbb{R}^d} q_{Z_\tau,Z_0}(z_\tau,z_{0}) dz_0 = \int_{\mathbb{R}^d} q_{Z_\tau|Z_{0}}(z_\tau|z_{0}) q_{Z_0}(z_{0}) dz_0, 
\end{equation}
where the conditional distribution $q_{Z_\tau|Z_{0}}(z_\tau|z_{0})$ is a Gaussian distribution $Z_\tau|Z_0 \sim \mathcal{N}(\alpha_{\tau}Z_0,\beta^2_{\tau}\mathbf{I}_d)$ as defined in Eq.~\eqref{eq:forward_diffusion_kernel},  and $q_{Z_0}(z_{0}) = p_{\Theta|Y}(\theta|y)$ is the target distribution.

{Substituting Eq.~\eqref{forward_states_pdf} and Eq.~\eqref{eq:bayes} into Eq.~\eqref{score_function}, we can re-write the score function as}
\begin{subequations}\label{score_function_approximation}
\begin{align}
S(z_{\tau},\tau)
&= \nabla_{z_\tau} \log\!\left(\int_{\mathbb{R}^d} q_{Z_\tau\mid Z_{0}}(z_\tau\mid z_{0})\, q_{Z_{0}}(z_{0})\, dz_0\right)
\label{score_function_approximation:a}\\[6pt]
&= \frac{\displaystyle \int_{\mathbb{R}^d}
\left(-\frac{z_\tau - \alpha_{\tau} z_0}{\beta^2_\tau}\right)
q_{Z_\tau\mid Z_{0}}(z_{\tau}\mid z_0)\,q_{Z_0}(z_0)\, dz_0}
{\displaystyle \int_{\mathbb{R}^d} q_{Z_\tau\mid Z_{0}}(z_\tau\mid \bar{z}_{0})\, q_{Z_{0}}(\bar{z}_{0})\, d\bar{z}_0}
\label{score_function_approximation:b}\\[6pt]
&= \int_{\mathbb{R}^d}
\left(-\frac{z_{\tau} - \alpha_\tau z_0}{\beta^2_\tau}\right)\,
w(z_{\tau}, z_0)\, dz_0.
\label{score_function_approximation:c}
\end{align}
\end{subequations}

{where the weight function $w(z_{\tau}, z_0)$ is defined as follows: recall that $q_{Z_0}(z_0) = p_{\Theta|Y}(\theta|y)$. Using Bayes’ rule}
$$
p_{\Theta\mid Y}(\theta\mid y)
= \frac{p_{Y\mid \Theta}(y\mid \theta)\, p_{\Theta}(\theta)}{\int_{\mathbb{R}^d} p_{Y\mid \Theta}(y\mid \bar\theta)\, p_{\Theta}(\bar\theta)\, d\bar\theta},
$$

we rewrite $q_{Z_0}(z_0)$ in  Eq.~\eqref{score_function_approximation:b} and define 
\begin{equation}\label{weight_function}
\begin{split}
    w(z_{\tau},z_{0}) &:= \frac{q_{Z_\tau|Z_0}(z_\tau|z_0) q_{Z_0}(z_0)} {\displaystyle \int_{\mathbb{R}^d}q_{Z_\tau|Z_0}(z_\tau|\bar{z}_0)q_{Z_0}(\bar{z}_0) d\bar{z}_0} 
    = \frac{q_{Z_{\tau}| \Theta,Y}(z_\tau|\theta,y)\, p_{Y|\Theta}(y|\theta)\,p_{\Theta}(\theta)} {\displaystyle  \int_{\mathbb{R}^d}\, q_{Z_{\tau}| \Theta,Y}(z_\tau|\bar{\theta}, y)\, p_{Y|\Theta}(y|\bar{\theta})\, p_{\Theta}(\bar{\theta}) d\bar{\theta}}
\end{split}
\end{equation}
where $\int_{\mathbb{R}^d} w(z_{\tau},z_0)\,dz_0 = 1$.


We can use the prior parameter samples to perform Monte Carlo estimation of the integral in Eq.~\eqref{score_function_approximation:c} and of the integral in the denominator of the weight function in Eq.~\eqref{weight_function}. Specifically, for any fixed observation $y$, the Monte Carlo estimator for the score function is
\begin{equation}\label{score_MC}
S(z_\tau,\tau) \approx S^{\rm MC}(z_\tau,\tau) := \sum^{N_{\rm prior}}_{n=1} - \frac{z_\tau - \alpha_\tau \theta^{(n)}}{\beta^2_\tau} w^{\rm MC}(z_\tau,\theta^{(n)}|y),
\end{equation}
with the weight function's approximation $w^{\rm MC}$ defined by
\begin{equation}\label{weight_MC}\scriptsize
 w^{\rm MC}(z_\tau,\theta^{(n)}|y) := \frac {\displaystyle\exp \left\{ \frac{-(z_\tau - \alpha_\tau \theta^{(n)})^2}{2\beta^2_\tau}\right\} \exp\left(- (y - {\mathcal{H}(u(x,t,\theta^{(n)})}))^\top\Sigma^{-1}(y - {\mathcal{H}(u(x,t,\theta^{(n)})}\right)}  {\displaystyle \sum_{n'=1}^{N_{\rm prior}} \exp \left\{ \frac{-(z_\tau - \alpha_\tau \theta^{(n')})^2}{2\beta^2_\tau}\right\} \exp\left(- (y - \mathcal{H}(u(x,t,\theta^{(n')}))^\top \Sigma^{-1}(y - \mathcal{H}(u(x,t,\theta^{(n')}))\right)},
\end{equation}
where $\Sigma \in \mathbb{R}^{q \times q}$ is the covariance matrix of the observation noise defined in the likelihood function in Eq.~\eqref{eq:likelihood}. Here, the solution $u$ may be defined by a low-fidelity model.

\paragraph{\bf\em {Step 2 (b): Constructing $\mathcal{S}_{\rm label}$}}
With the score estimator in Eq.~\eqref{score_MC}, the procedure for generating the labeled data set can be summarized as follows. {For $m = 1, \ldots, M$, we first generate one sample $y^{(m)}$ from $p_Y(y)$, which can be done by running another unconditional training-free diffusion model using the marginal sample set $\{y^{(n)}\}_{n=1}^{N_{\rm prior}}$ in Eq.~\eqref{eq:prior_sample}. }Then, we generate one sample $z^{(m)}$ from $\mathcal{N}(0, \mathbf{I}_d)$. Next, we set $Z_1 = z^{(m)}$ and $Y = y^{(m)}$ in Eq.~\eqref{eq:reverse_ODE} and solve the reverse ODE and define $z_0$ as the labeled data $\theta^{(m)}$ in Eq.~\eqref{eq:label_data}.  We summarize step 2 in Algorithm~\ref{alg:lf-alg}.





\paragraph{\bf \em Step 3: Supervised learning of the low-fidelity model using the labeled data}
Once $\mathcal{S}_{\rm label}$ is built, we train a fully-connected neural network model using the standard mean squared error loss to obtain the desired low-fidelity generative model $G^{\rm low}(y, z)$. This conditional generative model is capable of providing efficient posterior samples from the target distribution $p_{\Theta | Y}(\theta|y)$ for any given data observation of interest $Y=y$ from the distribution $p_Y(y)$. In practice, if the low-fidelity model provides satisfactory results for a specific parameter estimation problem, the user of this method does not need to use the high-fidelity model. Otherwise, they will use the low-fidelity model to build the high-fidelity model.

\subsection{Construction of the high-fidelity generative model}\label{sec:high}
If the prior data set $\mathcal{S}_{\rm prior}$ does not provide sufficiently dense coverage of the prior domain $\Gamma$ or if the solver is low-fidelity due to a reduced simulation runs, the low-fidelity generative model $G^{\rm low}$ may overestimate uncertainty when approximating the target conditional distribution $p_{\Theta|Y}(\theta|y)$ for a fix observation of interest $Y=y$. Hence, if a higher-fidelity approximation to the specific target conditional density $p_{\Theta|Y}(\theta|y)$ is desired, the following refinement procedure can be conducted to obtain the high-fidelity generative model $G^{\rm high}(z)$. The key idea is to use the distribution defined by $G^{\rm low}(y,z)$ for the fixed observation $y$ as a sampling distribution, and to refine the score estimation in Eq.~\eqref{score_MC} and Eq.~\eqref{weight_MC}. 
The procedure consists of the following steps. 
\begin{itemize}[leftmargin=20pt]  
    \setlength{\itemsep}{0.0cm}

    \item Use the low-fidelity generative model $G^{\rm low}(y,z)$ constructed in Section \ref{sec:low} to generate a set of refined prior samples, denoted by $\{\theta^{(k)}\}_{k=1}^K$, for the fixed $y$.

    %
    \item Perform kernel density estimation (KDE) based on the samples $\{\theta^{(k)}\}_{k=1}^K$ to obtain an approximation to the probability density function, denoted by $p^{\rm KDE}_{\Theta}(\theta)$ determined by $G^{\rm low}(y,z)$. {See comments on construction of this KDE in Section~\ref{sec:KDE_bandwidth}.} 
    
    \item Based on samples $\{\theta^{(k)}\}_{k=1}^K$ (where $K$ is typically taken to be quite high for for the KDE approximation and can be generated cheaply using $G^{\rm low}$), take a refined data set of size $N_{\rm refine} < K$ and solve the differential equation in Eq.~\eqref{eq:system} with a high-fidelity solver to obtain the new refined prior data set:
    \begin{equation}\label{eq:refined_prior}
      \mathcal{S}_{\rm prior} =  \{\theta^{(n)}, y^{(n)}\}_{n=1}^{N_{\rm refine}}. 
    \end{equation}

    Note that $y^{(n)}$ will be close to the fixed observation $y$ but not exactly equal to it due to the inaccuracy of the low-fidelity model. {See Section~\ref{sec:Nrefine_comment} for practical guidance on selecting $N_{\rm refine}$.}

\item Using $\mathcal{S}_{\rm prior}$, we now generate the  refined labeled data set, denoted by 
    \begin{equation}\label{eq:refined_labeled_data}
    \mathcal{S}_{\rm refine} =  \left\{\left(z^{(m)}, \theta^{(m)}\right): m = 1,\dots,M_{\rm refine}\right\},   
    \end{equation}
    by solving the reverse ODE for the fixed $y$ and using the updated estimators for the score function, i.e.,
    \begin{equation}\label{eq:HF_score}
     S^{\rm high}(z_\tau,\tau) := \sum^{N_{\rm refine}}_{n=1} - \frac{z_\tau - \alpha_\tau \theta^{(n)}}{\beta^2_\tau} w^{\rm high}(z_\tau,\theta^{(n)}|y),
\end{equation}
with the weight function's approximation $w^{\rm high}$ defined by
\begin{equation}\label{eq:HF_weight}
\resizebox{0.92\linewidth}{!}{$
 w^{\rm high}(z_\tau,\theta^{(n)}\!\mid\! y)
 = \frac{\exp\!\left\{-\tfrac12\!\left[\tfrac{(z_\tau-\alpha_\tau\theta^{(n)})^2}{\beta_\tau^2}
     + 2\,\mahn{\rN{n}}\right]\right\}
     \tfrac{p_{\Theta}(\theta^{(n)})}{p_{\Theta}^{\rm KDE}(\theta^{(n)})}}
    {\displaystyle\sum_{n'=1}^{N_{\rm refine}}
     \exp\!\left\{-\tfrac12\!\left[\tfrac{(z_\tau-\alpha_\tau\theta^{(n')})^2}{\beta_\tau^2}
     + 2\,\mahn{\rN{n'}}\right]\right\}
     \tfrac{p_{\Theta}(\theta^{(n')})}{p_{\Theta}^{\rm KDE}(\theta^{(n')})}}
$}
\end{equation}
where
\[
\rN{n} := y-\mathcal{H}\!\left(u(x,t,\theta^{(n)})\right). 
\]
Here $p_{\Theta}(\theta)$ is the prior probability density function and $p^{\rm KDE}_{\Theta}(\theta)$ is the kernel density estimation of the sample set $\{\theta^{(k)}\}_{k=1}^K$. Here $u$ is the solution obtained from a high-fidelity model. 

\item Use the new labeled data $\mathcal{S}_{\rm refine}$ to train another fully-connected neural network to obtain the high-fidelity generative model $G^{\rm high}(z)$. 
\end{itemize}

Once trained, the high-fidelity generative model $G^{\rm high}(z)$ is capable of efficiently mapping samples from the standard Gaussian distribution to the target distribution $p_{\Theta|Y}(\theta|y)$ for a fixed $y$ with improved accuracy over the low-fidelity model $G^{\rm low}(y,z)$. However, the improved accuracy is only valid for a specific observation $y$. The above procedure needs to be repeated if we need to obtain the high-fidelity model for another observation value. We summarize step 3 in Algorithm~\ref{alg:hf-alg}

\subsection{Comment on selection of $N_{\rm refine}$}\label{sec:Nrefine_comment}
 {To ensure the stability and accuracy of the high-fidelity generative model for the fixed observation $y$, a sufficient number of samples must be generated by the high-fidelity solver;} hence we propose a heuristic procedure for selecting $N_{\rm refine}$ when the target distribution is unavailable: select a refinement schedule such as $\{N^{(i)}_{\rm refine}\}_{i=1}^L$ of $L$ total candidates. Begin with an initial refined prior dataset $\mathcal{S}_{\rm prior}^{(1)}$ as in Eq.~\eqref{eq:refined_prior} of size $N^{(1)}_{\rm refine}$ and its corresponding labeled dataset $\mathcal{S}_{\rm refine}^{(1)}$ as in Equation ~\eqref{eq:refined_labeled_data}; form a KDE estimate ${p}_{1}^{\mathrm{HF}}$ using $\{\theta^{(m)}\}^{M_{\rm refine}}_{m=1}$. Next, solve the high-fidelity differential equation using the additional $\big(N_{\rm refine}^{(2)} - N_{\rm refine}^{(1)}\big)$ samples to obtain the larger prior data set $\mathcal{S}^{(2)}_{\rm prior} = \{(\theta^{(n)}, y^{(n)})\}_{n=1}^{N^{(2)}_{\rm refine}}$.  Note that additional samples are appended to previous datasets, and hence the datasets are nested $\mathcal{S}^{i-1}_{\rm prior}  \subset \mathcal{S}^{i}_{\rm prior} $.  For the fixed observation value $y$, use this dataset for estimation of the score in Eq.~\eqref{eq:HF_score},and run the reverse ODE process to obtain $\mathcal{S}^{(2)}_{\rm refine}$. Again, form a KDE estimate ${p}_{2}^{\mathrm{HF}}$ from the samples in $\mathcal{S}^{(2)}_{\rm refine}$; repeat this process until consecutive density approximations  ${p}_{i}^{\mathrm{HF}}$ and ${p}_{i+1}^{\mathrm{HF}}$ fall within a preselected threshold $\epsilon^{\rm TOL}$ of a discrepancy measure (e.g. Jensen–Shannon divergence). Once this tolerance is met, terminate the process with $N_{\rm refine} = N_{\rm refine}^{(i)}$ refined prior samples and $\mathcal{S}_{\rm refine} = \mathcal{S}^{(i)}_{\rm refine}$ labeled dataset. 
 

\subsection{Comment on KDE procedure}\label{sec:KDE_bandwidth}

{We now provide the details of the KDE procedures used in this work for estimating $p^{\rm KDE}_{\Theta}$ and $\{p_i^{\rm HF}\}^L_{i=1}$ defined in the previous section}. All kernel density estimates in the presented numerical examples are conducted using 10,000 generated samples to guarantee sufficient sample size for density estimation, which can be quickly synthesized as demonstrated in the numerical experiments. {We note that our numerical examples are restricted to 1D and 2D examples, hence the discussion pertains to these low-dimensional settings; future work for high dimensional parameter estimation tasks is ongoing work. }

The kernel density estimate influences two factors: (i) the sampling/proposal efficiency for constructing the high-fidelity generative model and (ii) the divergence values when benchmarking against a known ground-truth density. To address point (i), we note that while the efficiency of our method may benefit from a low-fidelity kernel density estimate $p^{\rm KDE}_{\Theta}$ which avoids over smoothing, determining such a KDE is not required for achieving computational gains over running the high-fidelity model alone, as demonstrated in our numerical examples. It suffices that the proposal KDE avoids severe over-smoothing and that its support covers that of the target density. Hence, we fix the KDE method to be SciPy's \texttt{gaussian\_kde} with Silverman's rule of thumb for bandwidth selection, given by $h = \left(\frac{4\hat{\sigma}^5}{3n}\right)^{1/5} \approx 1.06\hat{\sigma}n^{-1/5}$ where $\hat{\sigma}$ is the sample standard deviation and $n=K$ is the sample size, which has demonstrated good performance in practice.  


To address (ii), when an analytical reference density is available, we fit a Gaussian–kernel KDE to the 10{,}000 high-fidelity samples and {select the bandwidth that minimizes} the Kullback–Leibler divergence
between the true analytical distribution and the KDE over a logarithmically spaced grid of 100 candidates in \([10^{-4}, 1]\); the selected bandwidth is then kept fixed for all KDE estimates. This ensures that divergence comparisons across different approximations reflect true differences in sample quality rather than variations in bandwidth selection method. In practical settings, an analytical ground truth distribution is unavailable, hence this KDE estimation procedure is done only in the first example to demonstrate the accuracy of the method.

\begin{algorithm}[t]
  \caption{{Construction of $G^{\rm low}$}}
  \label{alg:lf-alg}
  \begin{algorithmic}[1]
    \Require  $\mathcal{S}_{\mathrm{prior}}$; labeled dataset size $M$; observation set $\{y^{(m)}\}^M_{m=1}$; proposal sample size $K$
    
    \Ensure $G^{\mathrm{\rm low}};  p^{\mathrm{KDE}}_{\Theta}$
    \State Using score estimator in Eq. (3.12), run reverse diffusion process to build low-fidelity labeled dataset $\mathcal{S}_{\rm label}$ of size $M$ 
    \State Train $G^{\mathrm{\rm low}}$ on $\mathcal{S}_{\mathrm{label}}$ 
    \State For fixed observation $y^{(m)}$, generate $\{\theta^{(k)}\}_{k=1}^{K}$ samples using $G^{\mathrm{low}}$
    \State Fit KDE $p^{\mathrm{KDE}}_{\Theta}$ to $\{\theta^{(k)}\}_{k=1}^{K}$
    
    \State \Return $G^{\mathrm{\rm low}};  p^{\mathrm{KDE}}_{\Theta}$
  \end{algorithmic}
\end{algorithm}

\begin{algorithm}[t]
  \caption{{Construction of $G^{\rm high}$}}
  \label{alg:hf-alg}
  \begin{algorithmic}[1]
    \Require  $G^{\rm low}$; $p^{\rm KDE}_{\Theta}$; labeled dataset size $M_{\rm refine}$; high-fidelity differential equation solver; refinement schedule $\{N^{(i)}_{\mathrm{refine}}\}_{i=1}^L$; fixed observation $y$; discrepancy measure $\mathcal{D}(\cdot | \cdot)$; tolerance $\varepsilon^{\mathrm{TOL}}$
    
    \Ensure $G^{\mathrm{high}}$

    \State Draw $N_{\rm refine}^{(1)}$ samples from the low-fidelity distribution $\theta^{(n)} \sim p_{\Theta}^{\rm KDE}$
    \State Solve with high-fidelity solver to build $\mathcal{S}_{\rm prior}^{(1)}$
    \State Use $\mathcal{S}_{\rm prior}^{(1)}$ and $p_{\Theta}^{\rm KDE}$ to estimate the score in Eq.~\eqref{eq:HF_score}, run the reverse diffusion process for fixed observation $y$ to build labeled dataset $\mathcal{S}^{(1)}_{\rm refine}$ of size $M_{\rm refine}$  
    \State Fit KDE $p_1^{HF}$ using $\theta \in \mathcal{S}^{(1)}_{\rm refine}$ . 
    \For{$2 \le i \le L$}
      \State Select $(N^{(i)}_{\mathrm{refine}} - N^{(i-1)}_{\mathrm{refine}})$ new samples from $p_{\Theta}^{\rm KDE}$ 
      
      \State Solve with the high-fidelity solver to obtain prior dataset $\mathcal{S}^{(i)}_{\rm prior}$
      \State Repeat line 3 using $\mathcal{S}_{\rm prior}^{(i)}$ and $p_{\Theta}^{\rm KDE}$ to obtain $\mathcal{S}^{(i)}_{\rm refine}$
      \State Fit KDE $ p^{\mathrm{HF}}_{i}$ using $\theta \in \mathcal{S}^{(i)}_{\mathrm{refine}}$  
      \If{ $\mathcal{D}\!\left( p^{\mathrm{HF}}_{i-1}\,\|\, p^{\mathrm{HF}}_{i}\right) < \varepsilon^{\mathrm{TOL}}$}
        \State \textbf{break}
      \EndIf

    \EndFor
    \State Set $\mathcal{S}_{\mathrm{prior}} = \mathcal{S}^{(i)}_{\mathrm{prior}}$ and $\mathcal{S}_{\mathrm{refine}} = \mathcal{S}^{(i)}_{\mathrm{refine}}$ 
    \State Train $G^{\mathrm{high}}$ on $\mathcal{S}_{\mathrm{refine}}$
    \State \Return $G^{\mathrm{high}}$
  \end{algorithmic}
\end{algorithm}

\section{Numerical Experiments}\label{sec:num}

In this section, we present numerical experiments to demonstrate the effectiveness of the proposed method. {Our method is compared against the true analytical posterior density when available, or against approximations obtained by Markov chain Monte Carlo (MCMC) and sequential Monte Carlo (SMC) sampling. For MCMC, we use the affine-invariant ensemble sampler implemented in the Python package \texttt{emcee} \url{https://emcee.readthedocs.io/en/stable} with 10 walkers and a burn-in of 20,000 steps. We drew 1,000 production steps per walker, yielding 10,000 posterior samples in total. For SMC sampling, we ran a tempered sampler with 10,000 particles across 50 temperature levels, resampling whenever the effective sample size (ESS) fell below fifty percent and applying one Metropolis-Hastings rejuvenation step per level. We additionally compare to a baseline procedure: the baseline consists of directly sampling from the prior distribution and conducting posterior density approximations via the generative procedure described in Step 2 with only the high-fidelity model. With this comparison, we highlight the computational/accuracy gains of conducting the two-step, bifidelity method over a one-step method in the presented examples.}

The example in Section~\ref{sec:ex1} benchmarks the
accuracy of our method for a problem with known analytical solution. {In Section~\ref{sec:ex2}, we estimate the viscosity parameter in viscous Burgers' equation, while in Section~\ref{sec:ex3} we conduct parameter estimation for the Lorenz 63 system. In Section~\ref{sec:ex4}, we estimate the drift and diffusion coefficients for a linear SDE. We conclude our examples in Section~\ref{sec:ex5} with an application in plasma physics, estimating parameters for a runaway electron simulation model.} Wall-clock times are computed by running our code on a server equipped with NVIDIA Tesla V100-PCIE with a GPU card of 32GB of memory.

\subsection{Illustrative Example with 1D parameter estimation}\label{sec:ex1}
In this subsection, we verify the accuracy of the proposed algorithm using an illustrative example with known analytical solution. The dynamical system and observation operator are defined as follows: \begin{equation} \mathcal{L}[u(\theta)] = u(\theta) - \theta^2 = 0, \quad \mathcal{H}(u(\theta)) = u, \quad \theta \in \Gamma = [-10, 10]. 
\end{equation} The noisy observation $y$ is expressed as: 
\begin{equation}\label{ex1:y}
y = \hat{\mathcal{H}}(u(\theta)) = \mathcal{H}(u(\theta)) + \epsilon = \theta^2 + \epsilon, 
\end{equation}
where the noise $\epsilon$ follows a Gaussian distribution, $\mathcal{N}(0, \sigma^2)$, with $\sigma^2 = 0.1$. The objective is to approximate the conditional distribution $p_{\Theta|Y}(\theta|y)$ for a given observation $y$. The exact conditional distribution is: \begin{equation}
p_{\Theta|Y}(\theta|y) = \frac{p_{Y|\Theta}(y|\theta)p_{\Theta}(\theta)}{p_{Y}(y)} = \frac{e^{-\frac{(y-\theta^2)^2}{2\sigma^2}} \mathbf{1}_{[-10,10]}(\theta)}{\int_{-\infty}^\infty e^{-\frac{(y-\theta^2)^2}{2\sigma^2}}d\theta}.
\end{equation}
This analytical solution allows us to assess the accuracy of the proposed method.

The dynamical system $\mathcal{L}$ has an explicit form that can be solved directly. Therefore, the main difference between low- and high-resolution simulations lies in the spatial discretization of samples from $\Gamma$. For training the low-fidelity diffusion model $G^{\rm low}(y,z)$ , we the first construct the prior dataset; we sample uniformly spaced points $\theta^{(n)}$ with a resolution of $\Delta \theta = 0.2$ across the domain $\Gamma = [-10, 10]$, yielding $N_{\rm prior} = 101$ samples. Using this data set, we simulate the reverse ODE in Eq.~\eqref{eq:reverse_ODE} using forward Euler and 500 timesteps from $\tau = 1$ to $\tau = 0$ to obtain labeled data for training the low-fidelity generative model $G^{\rm low}(y,z)$. 
Once trained, $G^{\rm low}$ can generate the low-fidelity density function over many $y \sim p_Y(y)$. To train the high-fidelity generative model $G^{\rm high}(z)$ for a specific observation $y$, we first use $G^{\rm low}$ to generate  samples $\{\theta^{(k)}\}_{k=1}^{K}$,
$K = 10,000$ for the fixed observation $y$. These samples are used to construct the proposal density function $p^{\rm KDE}_{\rm \Theta}$. 

Following the steps in Alg.~\ref{alg:hf-alg}, we use a refinement schedule for selecting the number of $N_{\rm refine}$ samples to draw from the proposal distribution; in this example, we start at 100 samples and increase by 50 samples up to 400 samples i.e.,
$\{N^{(i)}_{\mathrm{refine}}\}_{i=1}^{7} = \{100, \dots, 400\}$.
We take the discrepancy measure to be the Jensen-Shannon (JS) divergence in Eq.~\eqref{eq_js} below, and continue sampling according to the schedule until the divergence between consecutive KDE estimates falls below $10^{-2}$. We examine two test observations: $y=1$ and $y=9$.
Under this criterion, we obtain $N_{\mathrm{refine}} = 150$ for the $y=1$ case and $N_{\mathrm{refine}} = 200$ for the $y=9$ case. The values of the Jensen-Shannon divergence up to $N_{\rm refine} = 400$ are reported in Table~\ref{tab:js_divergence_comparison}.  We then construct the corresponding labeled dataset $\mathcal{S}_{\mathrm{refine}}$ for training of $G^{\rm high}$.
The refined dataset $\mathcal{S}_{\mathrm{refine}}$ is specific to the observation $y$.

\begin{figure}[H]
\centering
\includegraphics[width=\textwidth]{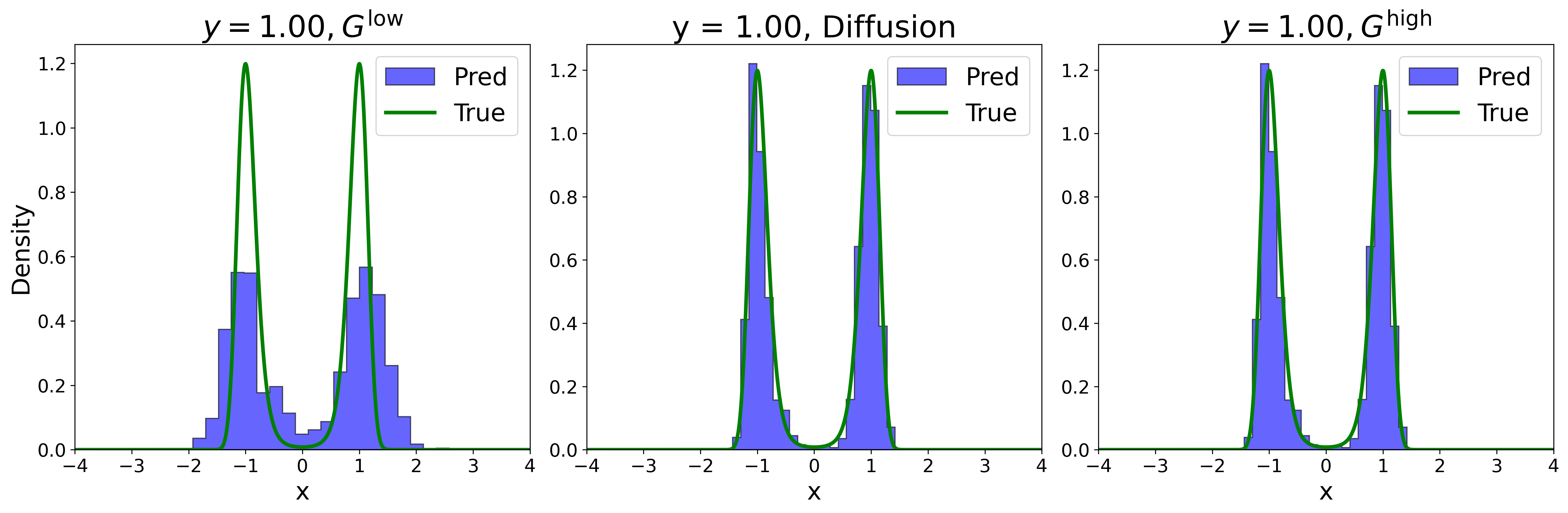}

\caption{{1D illustrative example}. The conditional distribution $p_{\Theta|Y}(\theta|y)$ for $y=1$. Left panel shows the density approximation obtained via the low-fidelity generative model $G^{\rm low}$. The KL divergence, computed according to Eq.~\eqref{eq_kl} is $0.26$. The middle panel displays the density approximation derived from labeled data by solving the reverse-time ODE model, yielding a KL divergence of 0.044. The right panel presents the density approximation generated by the high-fidelity generative model $G^{\rm high}$, which achieves also achieves a KL divergence of 0.045. The small KL divergence values for both the diffusion model and $G^{\rm high}$ indicate good approximation performance, with $G^{\rm high}$ learning the correct mapping from standard normal samples to posterior samples.
}
\label{fig:exact_setting_bimodal1}
\end{figure}

\begin{figure}[H]
\centering
\includegraphics[width=\textwidth]{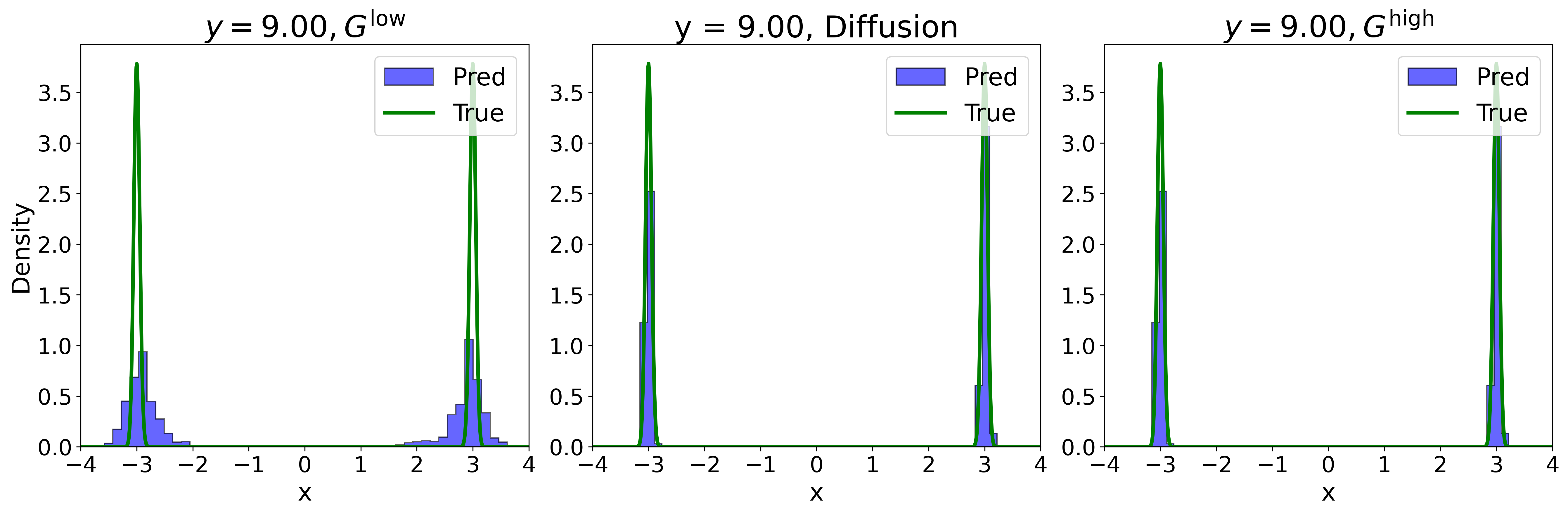}

\caption{{1D illustrative example}.The conditional distribution $p_{\Theta|Y}(\theta|y)$ for $y=9$. Left panel shows the density approximation obtained via the low-fidelity generative model $G^{\rm low}$. The KL divergence, computed according to Eq.~\eqref{eq_kl} is $1.09$. The middle panel displays the density approximation derived from labeled data by solving the reverse-time ODE model, yielding a KL divergence of 0.06. The right panel presents the density approximation generated by the high-fidelity generative model $G^{\rm high}$, which also achieves a KL divergence of 0.06. The small KL divergence values for both the ODE solution and $G^{\rm high}$ indicate good approximation performance, with $G^{\rm high}$.
}
 \label{fig:exact_setting_bimodal9}
\end{figure}

 Figs.~\ref{fig:exact_setting_bimodal1} and~\ref{fig:exact_setting_bimodal9} show the conditional distributions $p_{\Theta|Y}(\theta|y)$ for $y = 1$ and $y = 9$ respectively. {The left panels depict samples generated by the low-fidelity model $G^{\rm low}$, the middle columns are the samples generated in $\mathcal{S}_{\rm refine}$, and the right panels are samples generated from the high-fidelity generative model $G^{\rm high}$ for the specific data observation $y$.} The low-fidelity model $G^{\rm low}$ provides reasonable approximations, effectively capturing the regions of high posterior probability. The high-fidelity model $G^{\rm high}$, benefiting from the samples generated by the proposal distribution, achieves accurate approximations of the true distribution.
The KL divergence, used to quantify the difference between the exact distribution $p_{\rm exact}$ and the approximate distribution $p_{\rm approx}$, is defined as: 
\begin{equation}\label{eq_kl} 
D_{\rm KL}(p_{\rm exact} | p_{\rm approx}) = \int_{-\infty}^\infty p_{\rm exact}(x)\log\left(\frac{p_{\rm exact}(x)}{p_{\rm approx}(x)}\right)dx. 
\end{equation} 
The Jensen-Shannon divergence, used to determine the stopping criterion, is defined as:
\begin{equation}\label{eq_js}
\small
\mathrm{JS}\!\left(p_{\rm exact} | p_{\rm approx}\right)=\tfrac{1}{2}\,D_{\rm KL}\!\left(p_{\rm exact}\,\middle\|\,\tfrac{1}{2}\big(p_{\rm exact}+p_{\rm approx}\big)\right)+\tfrac{1}{2}\,D_{\rm KL}\!\left(p_{\rm approx}\,\middle\|\,\tfrac{1}{2}\big(p_{\rm exact}+p_{\rm approx}\big)\right).
\end{equation}

The integrals are approximated using a Riemann sum over a uniform mesh, specifically 1000 uniformly spaced points over [-4,4]. 

Table~\ref{tab:KL_bimodal} and Table~\ref{tab:KL_bimodal_y=9} compare the KL divergences of the true analytical distribution against the posterior from the two-step method, as well as against the posterior obtained by running the baseline procedure of using only samples from the prior distribution for the approximation. The results presented demonstrate that the two-stage method achieves lower KL with fewer solves, notably at smaller sample sizes.

\begin{table}[htbp]
\centering
\footnotesize
\begin{tabular}{lccccccccc}
\hline
\textbf{Method} & \textbf{150}\textsuperscript{*} & \textbf{200}\textsuperscript{*} & \textbf{250} & \textbf{300} & \textbf{350} & \textbf{400} \\
\hline
JS ($y=1^*$) & 0.0060 & 0.0028 & 0.0031 & 0.0026 & 0.0019 & 0.0008  \\
JS ($y=9^{**}$) &  0.0148 & 0.0026 & 0.0046 & 0.0032 & 0.00021 & 0.00013 \\
\hline

\end{tabular}
\caption{{Comparison} of Jensen–Shannon (JS) divergence across sample sizes for \(y=1\) and \(y=9\). Asterisks $(*)$ mark the sampling step at which the stopping criterion \(\varepsilon^{\mathrm{TOL}}=10^{-2}\) is met. The table shows that this tolerance indicates converging density estimates: beyond the marked number of samples, successive estimates continue to remain within a small JS discrepancy.}
\label{tab:js_divergence_comparison}
\end{table}

\begin{table}[H]
\centering
\footnotesize
\begin{tabular}{lccccccccc}
\hline
\textbf{Method (y=1)} & \textbf{100} & \textbf{150}\textsuperscript{*} & \textbf{200} & \textbf{250} & \textbf{300} & \textbf{350} & \textbf{400}  \\
\hline
Baseline KL & 0.2748 & 0.3507 & 0.3043 & 0.2059 & 0.1206 & 0.0693 & 0.0412  \\
$G^{\rm high}$ KL & 0.0588 & 0.0437 & 0.0512 & 0.0410 & 0.0417 & 0.0417 & 0.0325 \\
\hline
\end{tabular}
\caption{{Comparison} of KL divergence against the ground truth density across different sample sizes for observation $y=1$ for the baseline and bifidelity cases.}
\label{tab:KL_bimodal}
\end{table}

\begin{table}[H]
\centering
\footnotesize
\begin{tabular}{lccccccccc}
\hline
\textbf{Method (y=9)} & \textbf{100} & \textbf{150}\textsuperscript{**} & \textbf{200} & \textbf{250} & \textbf{300} & \textbf{350} & \textbf{400}  \\
\hline
Baseline KL & 0.2723 & 0.3489 & 0.3028 & 0.2041 & 0.1182 & 0.0385 & 0.0387  \\
$G^{\rm high}$ KL & 0.1293 &0.05064 & 0.0556 & 0.04938 & 0.0221 & 0.02293 & 0.0208 \\
\hline
\end{tabular}
\caption{{Comparison} of KL divergence against the ground truth density across different sample sizes for observation $y=9$ for the baseline and bifidelity cases.}
\label{tab:KL_bimodal_y=9}
\end{table}

\subsection{Viscous Burgers}\label{sec:ex2}

The solution to the viscous Burgers partial differential equation at equilibrium with perturbed boundary conditions is investigated, where $\theta$ represents the unknown viscosity parameter. The governing equations and observation operator are formulated as:
\begin{equation} \label{eq:viscous_burgers}
\begin{split}
\left\{
\begin{array}{l}
\mathcal{L}[u(x,t,\theta); \theta] = u_t(x,t,\theta) + u_x(x,t,\theta) - \theta u_{xx}(x,t,\theta) = 0, \quad \\[10pt]  
\mathcal{H}(u(x,t,\theta)) = [u(x_1,T,\theta), \dots, u(x_L,T,\theta)]^\top, \\[10pt]
x\in {\Omega} = [-1,1],  \ \ t\in[0,T], \ \ \theta \in \Gamma = [0.01,0.1],
\end{array}
\right.
\end{split}
\end{equation}
where $\Omega$ denotes the spatial domain, $\Gamma$ represents the admissible range of the viscosity parameter $\theta$, and $L$ denotes the number of sensor locations within spatial domain $\Omega$. The Dirichlet boundary conditions are prescribed as:
\begin{equation}
    u(-1,t) = 1 + \delta, \quad u(1,t) = -1, \quad t\in[0,T],
\end{equation}
where $\delta = 0.01$ represents the boundary perturbation magnitude. The observation operator maps the solution to discrete observations at $L$ sensor locations at time $T$. The noise-contaminated observations $\bm{y}$ are expressed as:

\begin{equation} \label{eq:noisy_data_viscous_burgers}
    \bm{y} = \hat{\mathcal{H}}(u(x,t,\theta)) = \mathcal{H}(u(x,t,\theta)) + \epsilon = [u(x_1,T,\theta) + \epsilon_1 \dots, u(x_L,T,\theta) + \epsilon_L]^\top.
\end{equation}

The measurement noise $\epsilon=[\epsilon_1,\ldots, \epsilon_L]^\top$ is characterized by a multivariate Gaussian distribution $\mathcal{N}(\mathbf{0}, \Sigma)$, with covariance matrix $\Sigma = \mathbf{I}\cdot\sigma^2 \in \mathbb{R}^{L \times L}$ and standard deviation $\sigma = 0.01$. Specifically, measurements are collected at $L = 21$ uniformly distributed sensor locations across the subdomain $[0,1]$ with spatial resolution $\Delta x = 0.05$. The objective is to approximate the posterior distribution  $p_{\Theta|Y}(\theta|\bm{y})$.

For the specified perturbation $\delta = 0.01$, a reference distribution is constructed to evaluate the accuracy of the proposed method. The solution of Eq.~\eqref{eq:viscous_burgers} at equilibrium for fixed $\theta$ is given by:
\begin{equation} \label{eq:exact_solution_viscous_burgers}
    u(x,\theta) = -A \tanh \left[\frac{A}{2\theta}(x-z_{\rm ex})\right],
\end{equation}
where $z_{\rm ex}$ denotes the location of the transition layer where $u(z_{\rm ex},\theta) = 0$ and $-A = \frac{\partial u}{\partial x}|_{x=z_{\rm ex}}$ is the slope at this location. 
This setting follows the super-sensitivity UQ problem studied in \cite{XiuK_IJNME04}. 


To construct a reference distribution conditioned on data observation of interest $\bm{y}$, we utilize MCMC by defining the following negative log-likelihood and assuming a uniform prior $p(\theta)$:

\begin{equation}{\label{eq:exact_likelihood_viscous_burgers}}
    -\log \mathcal{L}(\theta) = \frac{1}{2}[(\bm{y}-\mathcal{H}(u(x,t,\theta)))^\top\Sigma^{-1}(\bm{y}-\mathcal{H}(u(x,t,\theta)))], 
\end{equation}
where $\mathcal{H}$ is determined by Eq.\eqref{eq:exact_solution_viscous_burgers} at the $L$ sensor locations. We obtain $10,000$ samples from the posterior to perform KDE estimation. 

For numerical implementation, the Burgers equation is discretized using finite differences on a uniform grid comprising of $N_{\rm grid}$ points, where $N_{\rm grid}$ varies between low- and high-fidelity models. The temporal discretization is semi-implicit Euler method with time step $\Delta t = 0.01$. The simulation terminates after $5 \times 10^{4}$ time steps (upon reaching $T = 500$).
The low-fidelity diffusion model training data set $\mathcal{S}_{\rm prior} = \{\theta^{(n)}, \bm{y}^{(n)} \}_{n=1}^{N_{\rm prior}}$ is constructed where $\bm{y}^{(n)}$ follows Eq. \eqref{eq:noisy_data_viscous_burgers}. The initial parameter values $\theta^{(n)}$ are sampled uniformly with resolution $\Delta \theta = 9 \times 10^{-4}$ across $\Gamma = [0.01,0.1]$, yielding $N_{\rm prior} = 101$ samples. The corresponding solution values at time $T$ are obtained by first solving Eq.~\eqref{eq:viscous_burgers} using a spatial discretization of $N_{\rm grid}=400$ points. Measurement data are then collected at the $L=21$ sensor locations.  
The low-fidelity model $G^{\rm low}$ generates  samples $\{\theta^{(k)}\}_{k=1}^{K}$ for $K = 10,000$ to generate the probability density approximation.

\begin{figure}[h!]
    \centering
\includegraphics[width=\linewidth]{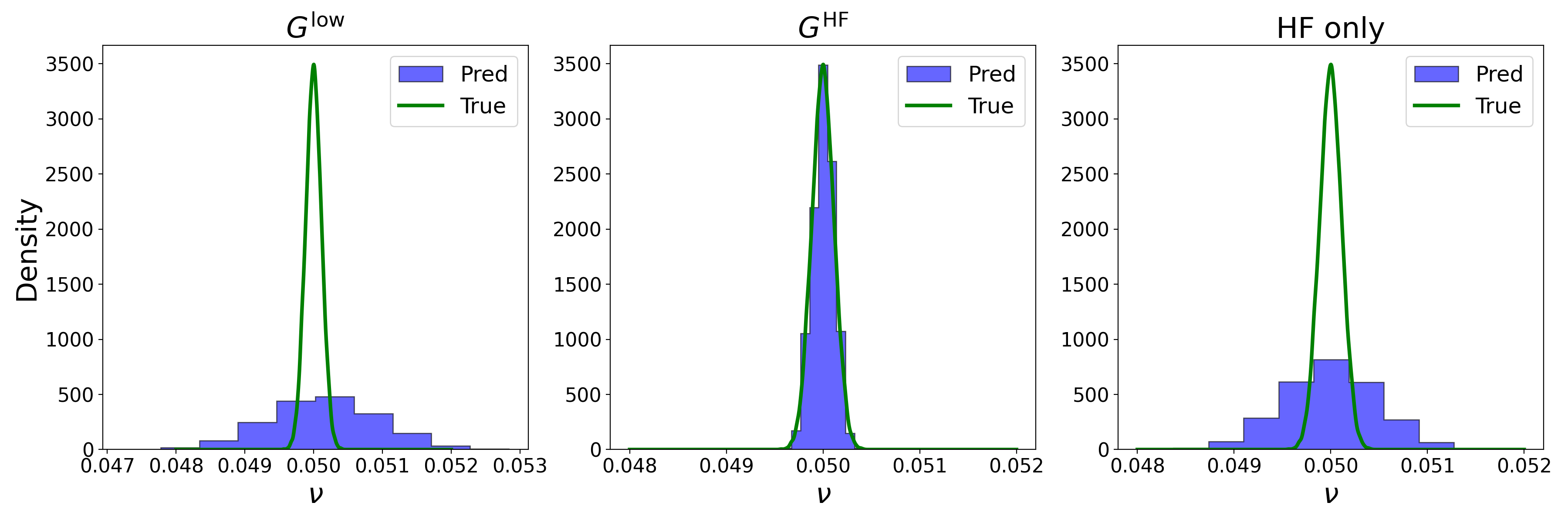}
\includegraphics[width=\linewidth]{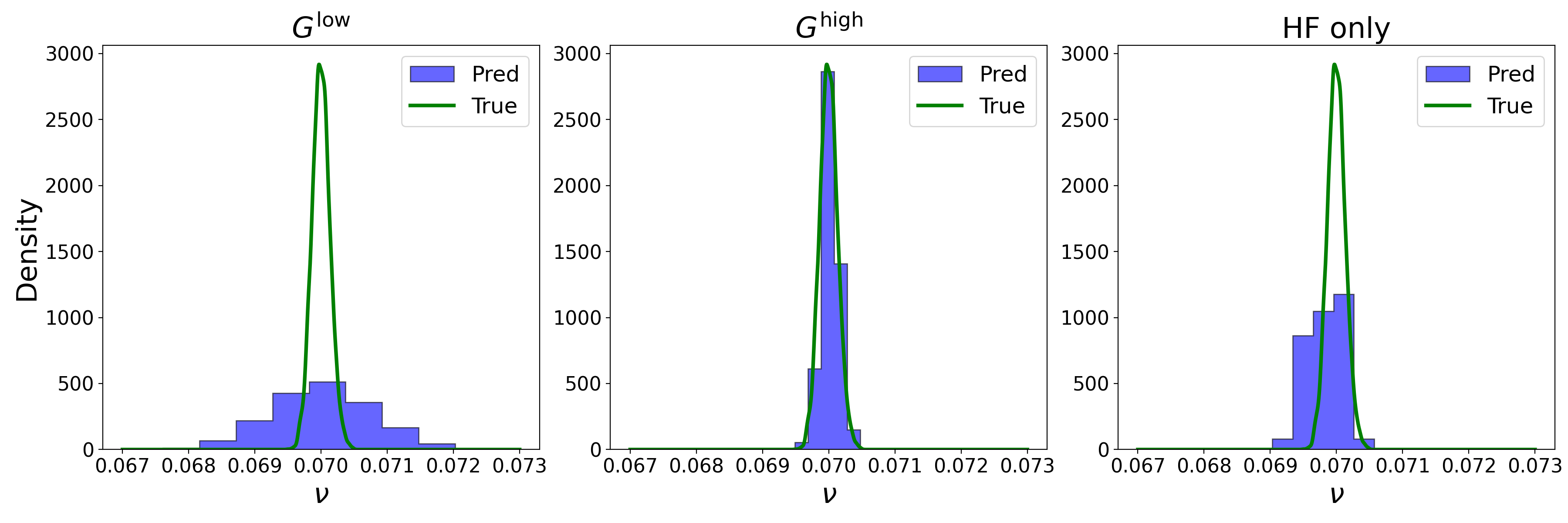}

\caption{{Viscous Burgers example}. The conditional distribution approximations for $p_{\Theta|Y}(\theta|\bm{y})$ for  $\bm{y}_{|\theta = 0.05}$ and $\bm{y}_{|\theta = 0.07}$. For observation $\bm{y}_{|\theta = 0.05}$ (top row): The left column shows the density approximation obtained via the low-fidelity generative model $G^{\rm low}$. The KL divergence for the low-fidelity approximation is $1.43$. The middle column presents the density approximation generated by the high-fidelity model $G^{\rm high}$ obtained using $N_{\rm refine} = 400$ samples, which achieves a KL divergence of $0.008$. The right column presents the baseline procedure again using 400 samples, taken over the prior domain, and its KL value is $0.96$. For observation $\bm{y}_{|\theta= 0.07}$ (bottom row): The left panel shows the density approximation from $G^{\rm low}$ with a KL divergence of $1.23$. The middle panel presents the $G^{\rm high}$ approximation with a KL divergence of $0.026$ using $N_{\rm refine} = 300$ samples, and the right panel presents the baseline procedure approximation with a KL divergence of $0.44$. The results demonstrate that $G^{\rm high}$ provide significantly more accurate approximations as evidenced by their substantially lower KL divergence values than $G^{\rm low}$ and the baseline.}
 \label{fig:viscous_burgers_example}
\end{figure}

The performance of both low- and high-fidelity generative models is evaluated using data ${\bm{y}}_{|\theta = 0.05}$ and ${\bm{y}}_{|\theta = 0.07}$, obtained by solving Eq.~\eqref{eq:viscous_burgers} using $N_{\rm grid} = 800$ with viscosity values $\theta = 0.05$ and $\theta = 0.07$ respectively. We again take the discrepancy measure to be the Jensen-Shannon divergence in Eq.~\eqref{eq_js}, and sample using increments of $100$ until the consecutive discrepancy value between KDEs fall below $\varepsilon^{\rm TOL} = 10^{-2}$.
Under this criterion, we obtain $N_{\mathrm{refine}} = 400$ for the $\theta= 0.05$ case and $N_{\mathrm{refine}} = 300$ for the $\theta= 0.07$ case. Results presented in Fig.~\ref{fig:viscous_burgers_example} demonstrate that the low-fidelity model provides an adequate initial approximation of the target density and identifies the relevant parameter domain. For both test cases, the Kullback-Leibler divergence between the reference and high-fidelity approximation distributions is of low order, indicating accurate approximation by the high-fidelity model. We compare to the baseline procedure using the same number of $N_{\rm refine}$ samples in each case, and can see the ability of the two-step method to more accurately capture the posterior distribution. 

Table.~\ref{tab:viscous_burgers_timing} presents the computational cost for each stage of the method, demonstrating the ability to efficiently sample from the posterior distribution $p_{\Theta|Y}(\theta|\bm{y})$ when compared to the runtime of MCMC. The results demonstrate the computational savings of our method while preserving accuracy when approximately sampling from the posterior density.
\renewcommand{\arraystretch}{1.25}
\begin{table}[h] 
\centering
\small
\begin{tabular}{lcccc}
\toprule
\textbf{Model} & \textbf{Data labeling} & \textbf{Training $G^{(\cdot)}$}  & \textbf{Synthesizing 10K samples} \\
\midrule
$G^{\rm low}$    & 0.23 seconds  & 1.59 seconds & 7.30e-04 seconds \\
$G^{\rm high}|\bm{y}_{|\theta = 0.05}$   & 0.40 seconds & 8.87 seconds & 4e-07 seconds \\
MCMC  & N/A  & N/A & 9.92 seconds \\
\midrule
$G^{\rm high}|\bm{y}_{|\theta = 0.07}$   & 0.56 seconds & 8.50 seconds & 4e-13 seconds \\
MCMC  & N/A  & N/A & 10.31 seconds \\
\bottomrule
\end{tabular}
\caption{Wall-clock run-time for different computational stages of the viscous Burgers example. Data labeling refers to running the reverse-time ODE for training data. Training $G^{(\cdot)}$ refers to the training time of the neural network model ($G^{\rm low}$ is trained for approximately 1300 epochs using early stopping with a validation split of 20\% and patience of 1000 epochs, while $G^{\rm high}$ models are trained for 10000 epochs). Synthesizing 10K samples refers to the run-time for neural network $G^{(\cdot)}$ to generate 10K samples. For $G^{\rm low}$, this is the average run time to synthesize the 10K samples for the two observations of interest. The results demonstrate the computational efficiency of the multi-level sampling approach for the viscous Burgers equation.}
\label{tab:viscous_burgers_timing}
\end{table}

\subsection{Lorenz 63 System}\label{sec:ex3}
Consider the three-dimensional Lorenz 63 system, which is known for exhibiting chaotic behavior. The system dynamics and observation operator are given by:
\begin{equation}
\left\{
\begin{aligned}
\mathcal{L}[u(t,\theta);\theta]
&=
\begin{bmatrix}
\dot{u}_1 - h(\gamma)(u_2 - u_1) \\
\dot{u}_2 - (u_1(\rho - u_3) - y_2) \\
\dot{u}_3 - (u_1u_2 - \beta u_3)
\end{bmatrix}
=
\begin{bmatrix}
0 \\
0 \\
0
\end{bmatrix},
\quad u(t_0)=u_0, \\[10pt]
\mathcal{H}(u(t,\theta))
&= u_1(t,\theta), \\[10pt]
t &\in [0,T], \quad \theta = (\gamma,\rho) \in \Gamma = [-5,5] \times [20,30].
\end{aligned}
\right.
\label{eq:chaotic_ode}
\end{equation}
where $\theta = (\gamma, \rho)$ represents the unknown parameters, and we fix $\beta = 2$. The observation operator $\mathcal{H}$ simulates the scenario in which only partial observation of the underlying system is feasible, i.e. for $u = [u_1,u_2,u_3]^\top \in \mathbb{R}^3$, we only observe solution trajectories of $u_1$. The function $h(\gamma) = \gamma^2$ relates to the parameter $\gamma$, and $\dot{u}$ denotes the time derivative. The initial condition is set to ${u}_0 = [-10,5,20]^\top$, and the time domain is defined as $T=1$.
The observations ${\bm y}$ consist of $(L+1)$ noisy measurements of the $u_1$ solution trajectory over time: 
\begin{equation} \label{eq:noisy_ODE_traj}
\bm{y} = \hat{\mathcal{H}}(u(t,\theta)) = \mathcal{H}(u(t,\theta)) + \mathbf{\epsilon} = [{u}_1(t_0;\theta) + \epsilon_0, \dots, {u}_1(t_{L};\theta) + \epsilon_{L}].
\end{equation}
The measurements are taken at uniform time intervals of $\Delta t = 0.02$, resulting in $(L+1) = 51$ observations over the time domain. The additive noise $\epsilon = [\epsilon_0, \epsilon_1, \dots, \epsilon_{L}]^\top$ is a vector of mean zero Gaussian random variables with covariance $\Sigma \in \mathbb{R}^{(L+1) \times (L+1)}$, $\Sigma = \mathbf{I} \cdot \sigma^2$ for $\mathbf{I} \in \mathbb{R}^{(L+1) \times (L+1)}$ with $\sigma^2 = 0.1$.  The goal is to approximate the two-dimensional posterior distribution $p_{\Theta|Y}(\theta|\bm{y})$ for parameters $\theta = (\gamma,\rho)$ given data observation $\bm{y}$. 


For the bifidelity diffusion model training, we employ different spatial resolutions for the parameter space $\theta = (\gamma, \rho)$. For the low-fidelity generative model $G^{\rm low}(y,z)$, we construct the training data set $\mathcal{S}_{\rm prior} = \{\theta^{(n)}, \bm{y}^{(n)} \}_{n=1}^{N_{\rm prior}}$ by sampling the $21 \times 21$ uniform grid over the parameter domain $\Gamma = [-5,5] \times [20,30]$ with spatial resolution $\Delta\gamma = \Delta \rho = 0.50$, yielding $N_{\rm prior} = 441$ samples. Using this data set and the reverse ODE simulation, we obtain the labeled training data for $G^{\rm low}$. The trained $G^{\rm low}$ is then used to generate the data set $\{\theta^{(k)}\}_{k=1}^K$ with $K = 10,000$ to approximate $p^{\rm KDE}_{\Theta}(\theta)$.

 The refined prior dataset is constructed by sampling $N_{\mathrm{refine}}$ parameter values from the low-fidelity distribution. We use consider a sampling refinement schedule defined by increments of 1000 samples. To ensure convergence of the KDE estimate, we require that two iterations of consecutive KDEs have Jensen–Shannon divergence below the specified tolerance, i.e., $\mathcal{D}\big(p_{i}^{\mathrm{HF}}|p_{i+1}^{\mathrm{HF}}\big)<\varepsilon_{\mathrm{tol}}$ for $i$ and $i+1$. Here we set the tolerance to $\varepsilon_{\mathrm{tol}}=10^{-2}$.

\begin{figure}[h!]
\centering
\includegraphics[width=\textwidth]{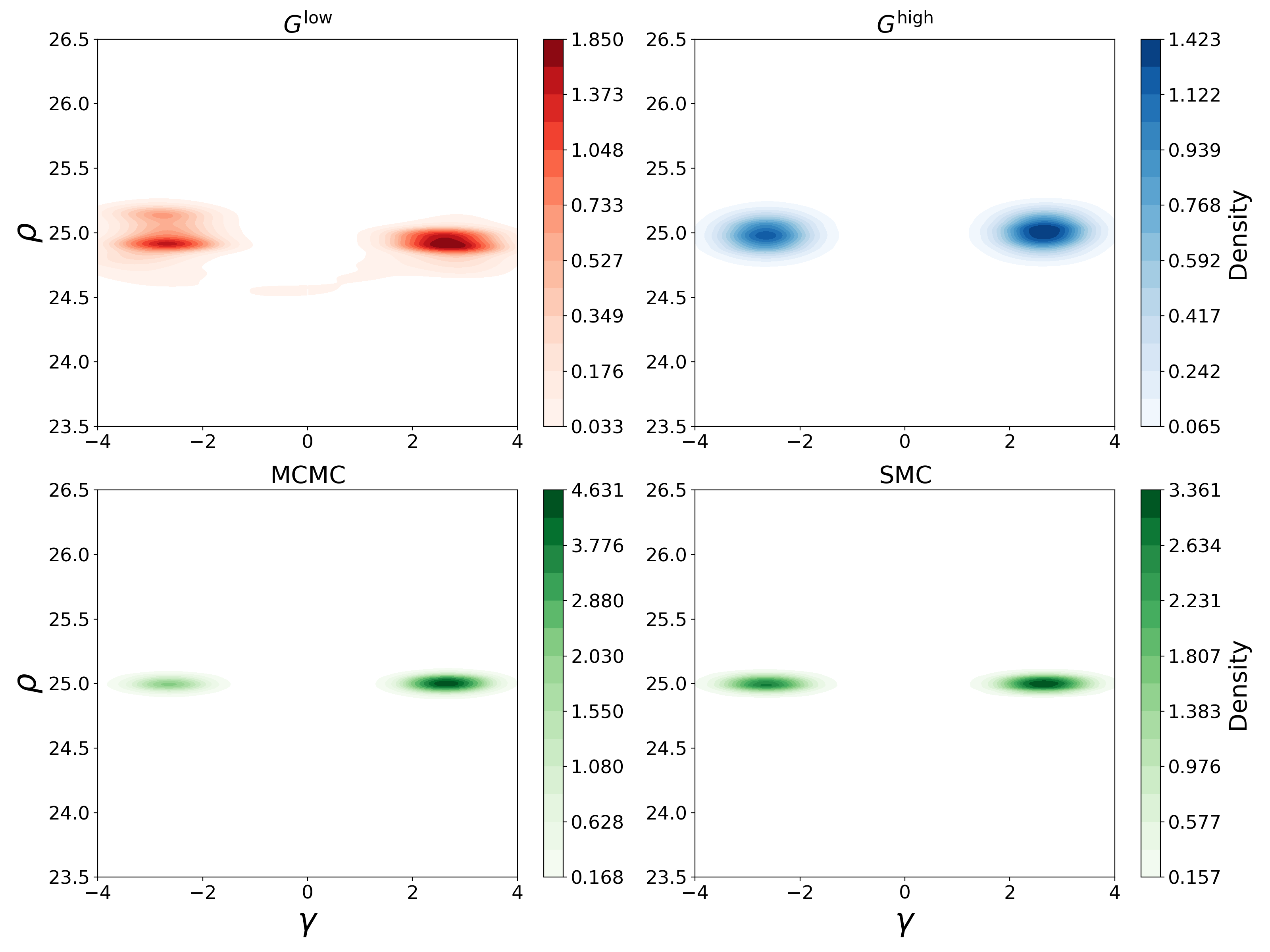}
\caption{{Chaotic ODE example}. Density approximations for observation $\bm{y}_1$, generated using parameters $(\gamma,\rho) = (\sqrt{7},25)$. The bimodal structure centers at $(-\sqrt{7},25)$ and $(\sqrt{7},25)$, reflecting the symmetry of the Lorenz system where both $(\gamma,\rho)$ and $(-\gamma,\rho)$ produce equivalent dynamics. Top left: Density approximation obtained via the low-fidelity generative model $G^{\rm low}(y,z)$, trained using $21 \times 21$ parameter pairs over domain $D = [-5,5] \times [20,30]$ (441 total samples), showing clear identification of the bimodal structure. Top right: High-fidelity density approximation generated by $G^{\rm high}(z)$, exhibiting balanced and well-separated modes, capturing the system's inherent symmetry. Bottom left: MCMC results using 10 walkers within the refined domain 
with 20,000 burn-in steps, showing less balanced representation of the bimodal structure compared to our proposed method. Bottom right: Sequential Monte Carlo with 10,000 particles identifies both modes with improved balance of modes.}
\label{fig:chaotic_ODE_1}
\end{figure}
\begin{figure}[h!]
\centering
\includegraphics[width=\textwidth]{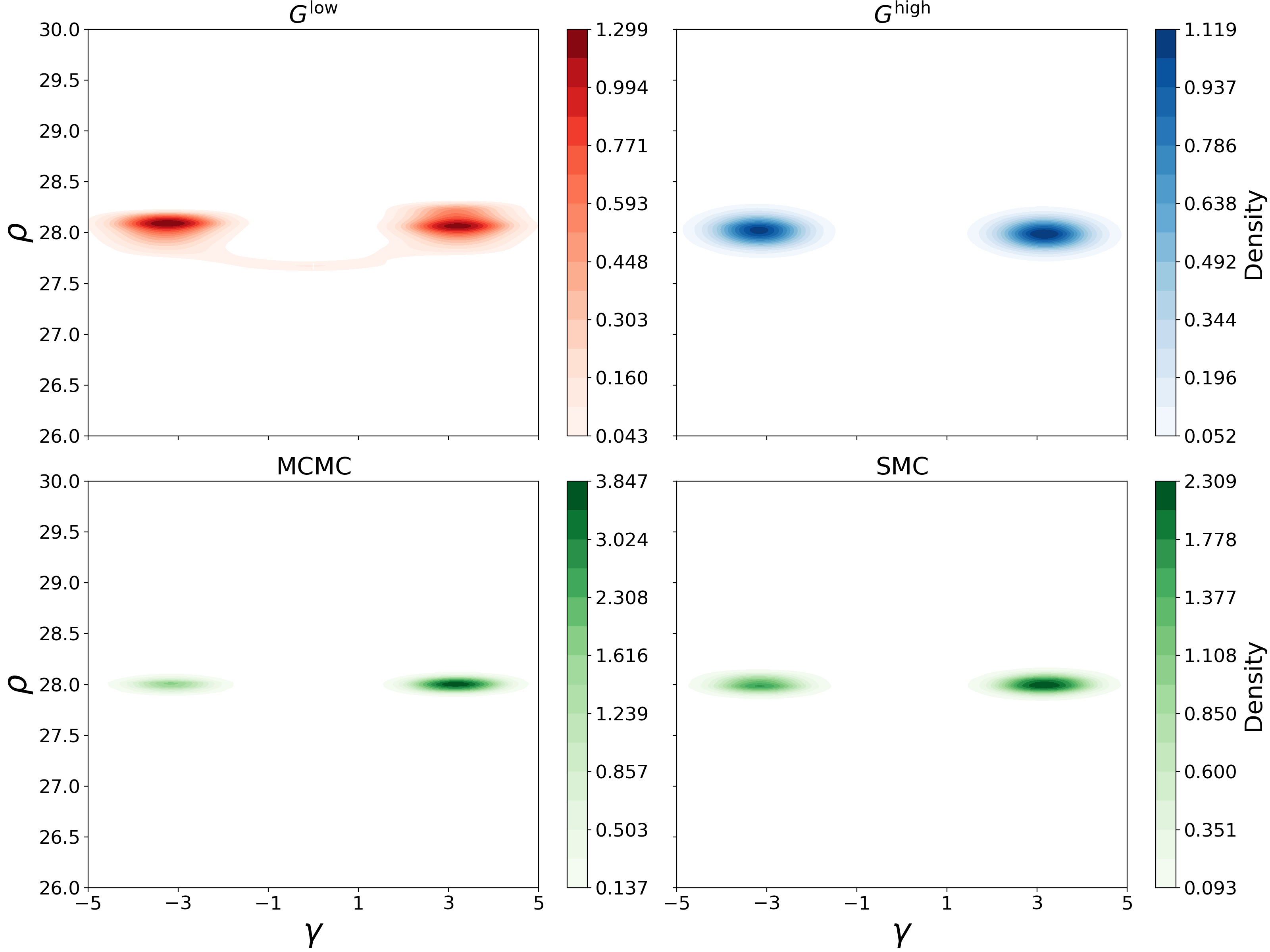}

\caption{{Chaotic ODE example}. Density approximations for observation $\bm{y}_2$, generated using parameters $(\gamma,\rho) = (\sqrt{10},28)$. The bimodal structure centers at $(-\sqrt{10},28)$ and $(\sqrt{10},28)$, reflecting the symmetry of the Lorenz system where both $(\gamma,\rho)$ and $(-\gamma,\rho)$ produce equivalent dynamics. Top left: Density approximation obtained via generative model $G^{\rm low}(y,z)$, trained using $21 \times 21$ parameter pairs $\theta = (\gamma,\rho)$ over domain $D = [-5,5] \times [20,30]$ (441 total samples). Top right: High-fidelity density approximation generated by $G^{\rm high}(z)$, exhibiting balanced and well-separated modes, capturing the system's inherent symmetry. Bottom left: MCMC results using 10 walkers within the refined domain 
with 20,000 burn-in steps, showing less balanced representation of the bimodal structure compared to our proposed method. Bottom right: Sequential Monte Carlo with 10,000 particles identifies both modes with improved balance of modes and slight imbalance.}
\label{fig:chaotic_ODE_2}
\end{figure}

We evaluate the low- and high-fidelity generative models using two different observations $\bm{y}_1$ and $\bm{y}_2$, obtained by solving Eq.~\eqref{eq:chaotic_ode} with parameters $(\gamma, \rho) = (\sqrt{7},25)$ and $(\gamma, \rho) = (\sqrt{10},28)$, respectively. The expected bimodal behavior centered at $(\pm\gamma, \rho)$ in this system arises from its symmetry properties. This symmetry, combined with the chaotic nature of the system, naturally leads to two equally probable parameter regions that can produce similar observations, resulting in an evenly distributed bimodal posterior distribution. 
In Figs.~\ref{fig:chaotic_ODE_1} and~\ref{fig:chaotic_ODE_2}, we compare results to MCMC as well as sequential Monte Carlo sampling. We compare these methods to the posterior distribution formed by generating 10,000 samples via the trained high-fidelity generative model $G^{\rm high}$.  It is observed that while MCMC fails to capture the symmetric bimodal structure for $y$, the score-based diffusion model reveals this fundamental characteristic of the posterior distribution. We see good agreement when comparing our method results to the SMC method, however our method benefits from lower computational run-times; The comparative wall-clock run-times for the first test case are recorded in Table~\ref{tab::chaotic_ode_table}.

\begin{figure}[h!]
\centering
\includegraphics[width=\textwidth]{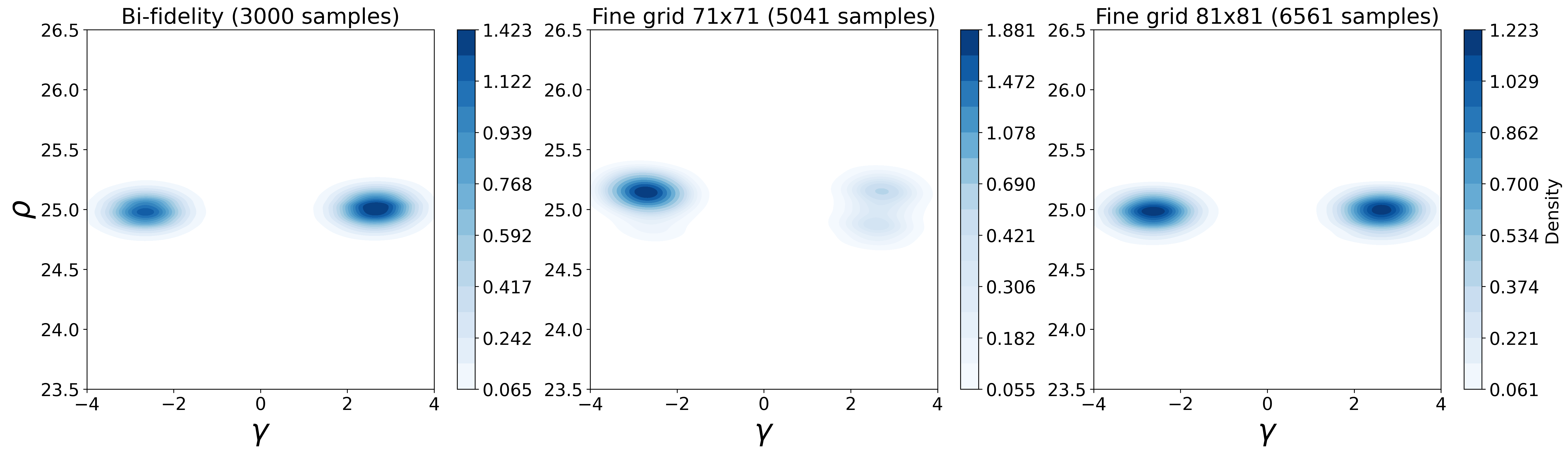}

\caption{{Chaotic ODE example}. Left: Density approximation from our proposed bifidelity method; (3,000 samples) denotes the number of high-fidelity solves used to obtain the posterior density. 
Middle: Posterior distribution from defining the prior dataset on a $61\times 61$ grid over $[-5,5]\times[10,20]$; the imbalance between the two modes and apparent over-smoothing highlights the advantage of the bifidelity approach under large, uninformative priors. 
Right: Posterior distribution from an $81\times 81$ grid over the prior domain, which agrees well with the posterior obtained by our bifidelity method.
}
\label{fig:chaotic_ODE_comp}
\end{figure}

Additionally, in Figure~\ref{fig:chaotic_ODE_comp}, we compare the bifidelity density with the result of the baseline procedure; specifically, we define the prior dataset on a $61\times 61$ grid (3,721 samples) over the prior domain and generate labeled data for the $\bm{y}_1$ case. Despite using more samples, the bifidelity posterior fails to capture the second mode. The large, uninformative prior domain requires a finer mesh to resolve the posterior; therefore, we also compare against an $81\times 81$ grid (6,561 samples), which agrees well with our posterior and highlights the potential computational savings of the bifidelity method.

\renewcommand{\arraystretch}{1.25} 
\begin{table}[H] 
\centering
\small
\begin{tabular}{lccc}
\toprule
\textbf{Model} & \textbf{Data labeling} & \textbf{Training $G^{(\cdot)}$} & \textbf{Synthesizing 10K samples} \\
\midrule
$G^{\rm low}$    & 0.53  seconds  & 4.72 seconds & 8.42e-04 seconds \\
\midrule
$G^{\rm high}|\bm{y}_1$   & 22.30 seconds & 7.24 seconds & 4.27e-04 seconds \\
$\text{MCMC}$   & N/A & N/A & 105.10 seconds \\
$\text{SMC}$   & N/A & N/A & 100.43 seconds \\
\bottomrule
\end{tabular}
\caption{Wall-clock run-time for different computational stages of the chaotic ODE (Lorenz 63) example. Data labeling refers to running the reverse-time ODE for training data. 
Training $G^{(\cdot)}$ refers to the training time of the neural network model (taken to be 3 hidden layers, 100 neurons per layer, trained for 4000 epochs). For data labeling, we used $N_{\rm refine} = 3000$ samples for observation $\bm{y}_1$. Synthesizing 10K samples refers to the run-time for neural network $G^{(\cdot)}$ (MCMC, SMC) to generate 10K samples. The results demonstrate the computational efficiency of generating parameter samples for the chaotic ODE system via the bifidelity, score-based diffusion methods when compared to standard sampling methods.}
\label{tab::chaotic_ode_table}
\end{table}

\subsection{Linear SDE}\label{sec:ex4} \label{subsec::SDE_linear}

In this section, we consider the one dimensional Ornstein–Uhlenbeck process defined as, 
\begin{equation}\label{eq:SDE_example_eqn}
\left\{
\begin{aligned}
\mathcal{L}[Y_t(\theta);\theta]
&= dY_t - (h(\mu) - Y_t)\,dt - g(\sigma)\,dW_t = 0,
\quad Y_0 = y, \\[10pt]
t &\in [0,T], \quad
\theta = (\mu,\sigma) \in \Gamma = [-10,10] \times [-10,10].
\end{aligned}
\right.
\end{equation}
where $T=1$ and $\theta = (\mu,\sigma)$ are unknown parameters of interest.  We define $h$ as a function of unknown parameter $\mu$ and $g$ as a function of unknown parameter $\sigma$. Specifically we fix $h(\mu) = \mu^2$ and $g(\sigma) = \sigma^2$ and the initial condition as $Y_0 = 1.5$.  The observation operator $\mathcal{H}$ is defined by the statistics of solution $Y$ at time $T$, i.e.,
\begin{equation}\label{eq:forward_model_sde}
\bm{y} = \mathcal{H}(Y_t(\theta)) = 
[\mathbb{E}[Y_T|\theta],
 S[Y_T|\theta]]^\top
\end{equation}
where $\mathbb{E}[\cdot|\theta]$ denotes the conditional expectation given parameter $\theta$ and $S[\cdot|\theta]$ refers to the conditional standard deviation.

To numerically approximate the observation $\bm y$, we consider the Monte-Carlo approximation of $\mathcal{H}$. Using the Euler–Maruyama method with $N_{\rm MC}$ simulation paths, we estimate the statistics in Eq.~\eqref{eq:forward_model_sde} for fixed $\theta$, where $N_{\rm MC}$ is an adjustable hyper-parameter controlling the accuracy of the simulation. The Monte Carlo approximation is given by:
\begin{equation} \label{SDE_sample_mean_sample_var}
\mathcal{H}^{\rm MC}(Y_t(\theta)) = \begin{bmatrix}
\tilde{\mathbb{E}}[Y_T|\theta], \
\tilde{S}[Y_T|\theta]
\end{bmatrix}^\top = \begin{bmatrix}
\frac{\sum_{i=1}^{N_{\rm MC}} Y^{(i)}_T(\theta) }{N_{\rm MC}}, \ \
\sqrt{ \frac{\sum_{i=1}^{N_{\rm MC}}\left(Y^{(i)}_T(\theta) - \tilde{\mathbb{E}}[Y_T|\theta]\right)^2}{N_{\rm MC} - 1}}
\end{bmatrix}^\top,
\end{equation}
where $Y^{(i)}_T(\theta)$ denotes the position of the $i$-th trajectory at time $T$ for fixed $\theta$. The output data from the simulation model can then be expressed as:
\begin{equation}\label{error_MC_SDE}
   {\bm{y}} = \mathcal{H}^{\rm MC}(Y_t(\theta)) = \mathcal{H}(Y_t(\theta)) + \epsilon,
\end{equation}
where $\epsilon \sim \mathcal{N}(0, \Sigma)$ represents the Monte Carlo estimation error. To understand the uncertainty in our parameter estimates due to this model error, we approximate the covariance matrix $\Sigma$ as:
\begin{equation} 
\scriptsize
\label{variance_estimator_SDE_example}
{\Sigma} \approx \hat{\Sigma} = \begin{bmatrix}
\frac{1}{N_{\rm grid}} \sum\limits_{i=1}^{N_{\rm grid}} \left( \tilde{\mathbb{E}}[Y_T|\theta] - {\mathbb{E}}[Y_T|\theta] \right)^2 & 0 \\[10pt]
0 & \frac{1}{N_{\rm grid}} \sum\limits_{i=1}^{N_{\rm grid}} \left(\tilde{S}[Y_T|\theta] - {S}[Y_T|\theta] \right)^2
\end{bmatrix},
\end{equation}
where $N_{\rm grid} = 20$ points are randomly sampled from the parameter domain $\Gamma$. In this example we assume no access to the exact form of the SDE in \eqref{eq:SDE_example_eqn}, we use $N_{\rm MC} = 10^5$ simulations to obtain the ground truth statistics in Eq.~\eqref{variance_estimator_SDE_example}.

The goal is to approximate the posterior density $p_{\Theta|Y}(\theta|\bm{y})$ for a given observation $\bm{y}$. While the expectation $\mathbb{E}[\cdot]$ and standard deviation $S[\cdot]$ of the solution $Y_T$ of Eq.~\eqref{eq:SDE_example_eqn}  can be computed exactly for known $\theta$, our data-driven setting assumes the forward model is unknown. Thus, we rely solely on numerical simulation to generate input-output pairs for parameter estimation.

The generation of training data for both low- and high-fidelity generative models differs in two aspects: the resolution of the parameter domain $\Gamma$ and the accuracy of the Euler–Maruyama method controlled by $N_{\rm MC}$. For the low-fidelity generative model $G^{\rm low}$, we construct the training data set $\mathcal{S}_{\rm prior} = \{\theta^{(n)}, \bm{y}^{(n)}\}_{n=1}^{N_{\rm prior}}$ by sampling a uniform $21 \times 21$ grid over $\Theta = [-10,10] \times [-10,10]$ with mesh size $\Delta \mu = \Delta \sigma = 1$, yielding $N_{\rm prior} = 441$ samples. We then generate corresponding outputs $\bm{y}^{(n)}$ using the Euler–Maruyama method with $N_{\rm MC} = 2000$ paths in Eq.~\eqref{SDE_sample_mean_sample_var}.
For a given observation $\bm{y}$, we construct the kernel density estimate from samples generated by $G^{\rm low}$. In this example, we employ a refinement schedule with increments of $500$ samples drawn from the low-fidelity distribution. A tolerance of $\varepsilon^{\mathrm{TOL}} = 10^{-2}$ is used for the Jensen--Shannon divergence between consecutive KDE constructions, requiring criteria to be satisfied over two successive iterations. The criterion is satisfied at \(N_{\mathrm{refine}} = 1500\) for \(\bm{y}_1\) and \(N_{\mathrm{refine}} = 3000\) for \(\bm{y}_2\).
The high-fidelity solution data are generated using $N_{\rm MC} = 10000$ paths in Eq.~\eqref{SDE_sample_mean_sample_var}, and the resulting refined prior data with $N_{\rm refine}$ samples is used to construct the labeled data set $\mathcal{S}_{\rm refine}$.

\begin{figure}[h!]
\centering
\includegraphics[width=\textwidth]{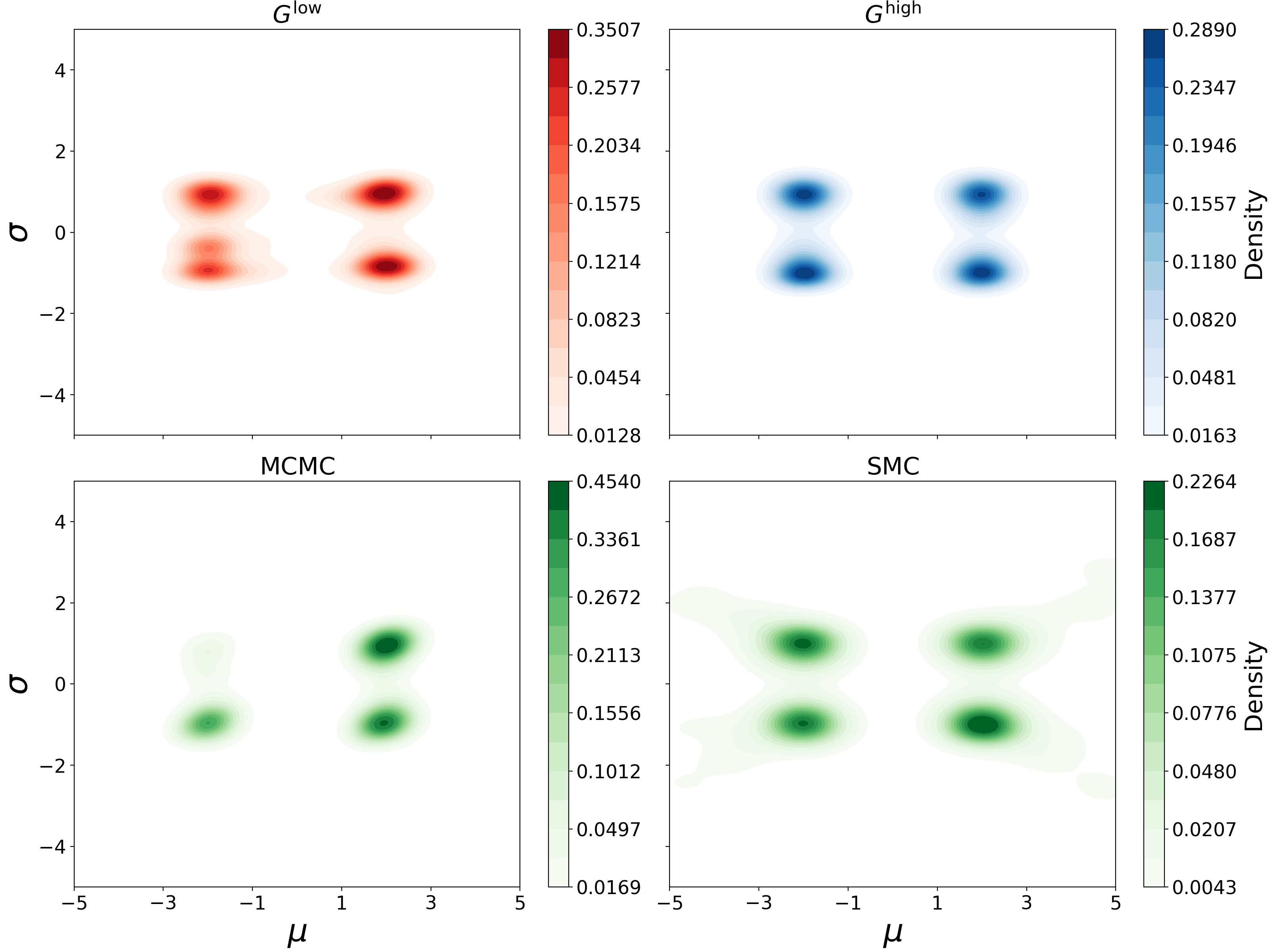}

\caption{{Linear SDE example}. Parameter estimation results for observation $\bm{y}_1$ corresponding to true parameters $(\mu,\sigma) = (2,1)$. The four panels demonstrate the accuracy of our approach compared to MCMC and SMC sampling: Top right: Initial low-fidelity generative model $G^{\rm low}(y,z)$. Top left:  Density approximation defined by high-fidelity model $G^{\rm high}(z)$.  Bottom left: MCMC sampling. Bottom right: SMC sampling. Our method successfully captures the expected four equal modes centered at $(\pm\mu,\pm\sigma)$, while MCMC struggles to identify this symmetric structure in the posterior distribution. SMC captures the four modes and appears to over-smooth, but displays good agreement with our results.}
\label{fig:SDE_1}
\end{figure}

\begin{figure}[h!]
\centering
\includegraphics[width=\textwidth]{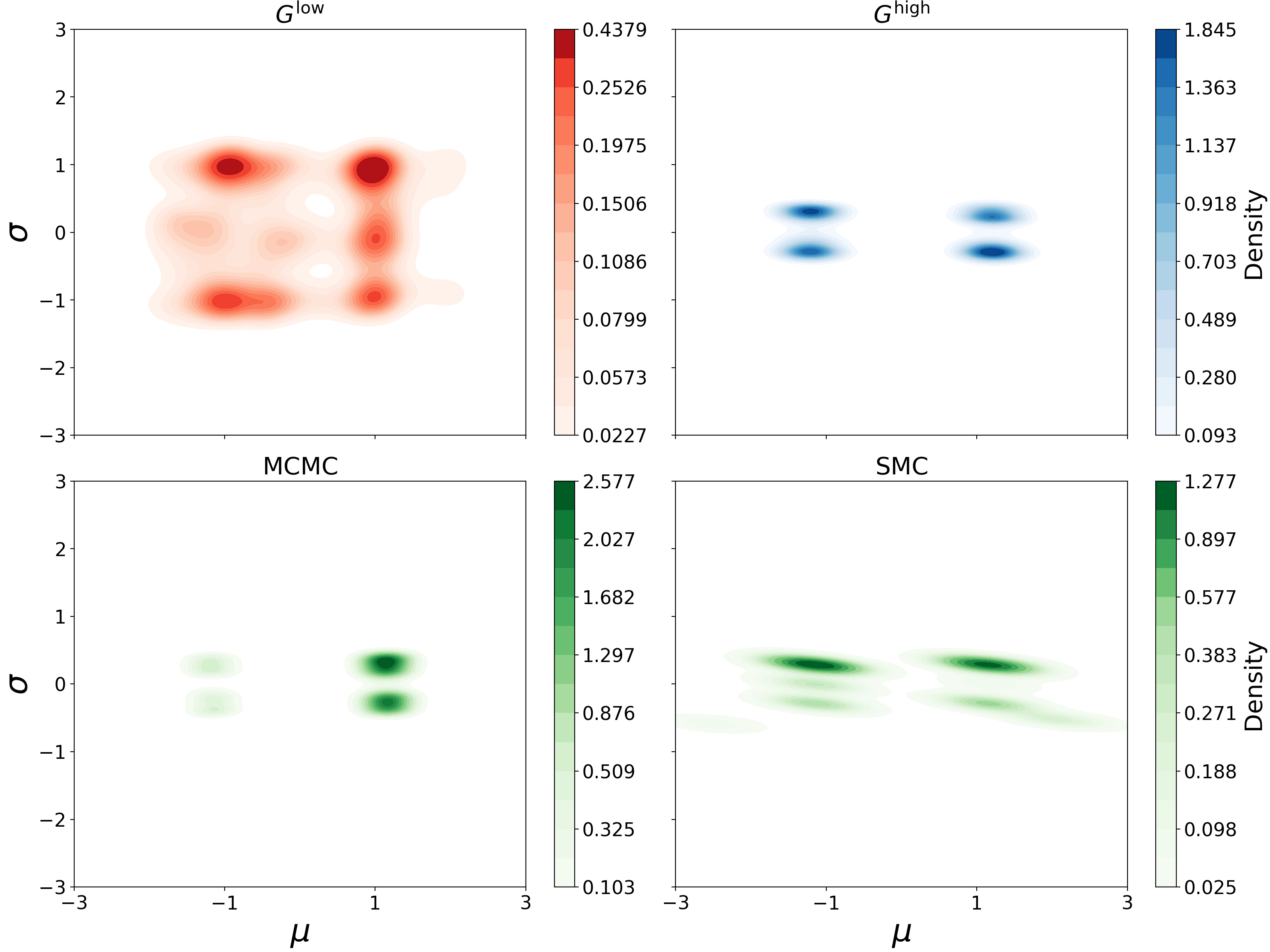}

\caption{{Linear SDE example}. Parameter estimation results for observation $\bm{y}_2$ corresponding to true parameters $(\mu,\sigma) = (1.2,0.3)$. Top right: Initial low-fidelity generative model $G^{\rm low}(y,z)$. Top left:  Density approximation defined by high-fidelity model $G^{\rm high}(z)$.  Bottom left: MCMC sampling. Bottom right: SMC sampling. While our method successfully captures the expected four equal modes centered at $(\pm\mu,\pm\sigma)$, MCMC and SMC struggle to identify the symmetric structure in the posterior distribution.}
\label{fig:SDE_2}
\end{figure}

\begin{figure}[h!]
\centering
\includegraphics[width=\textwidth]{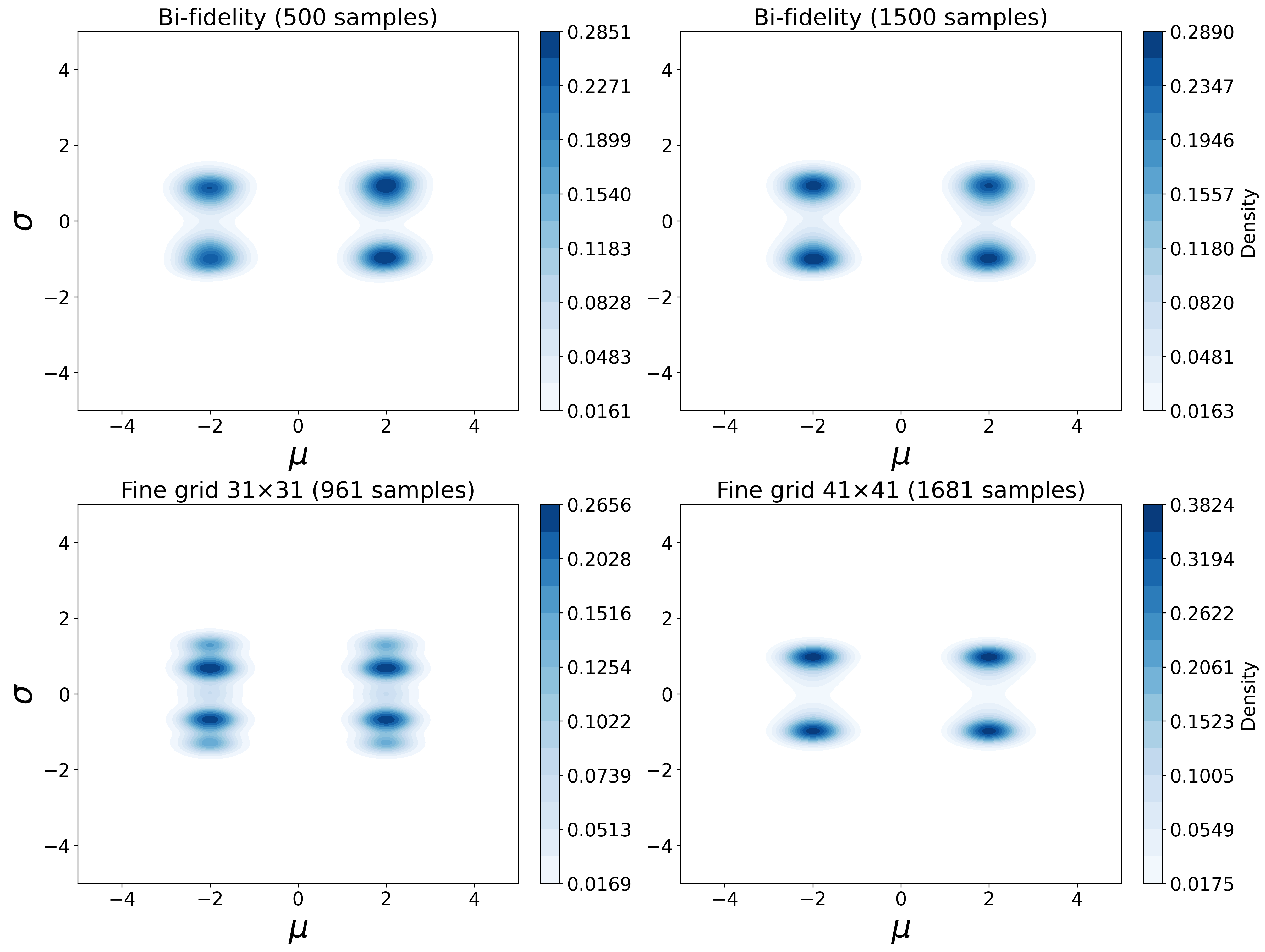}

\caption{{Linear SDE example}. Top left: Density approximation from our proposed bifidelity method using only 500 high-fidelity samples with $N_{\rm MC} = 10,000$ used to obtain solution data. 
Top right: Density approximation from our proposed bifidelity method using 1500 high-fidelity samples. Bottom left: Posterior distribution from defining the prior dataset on a $31\times 31$ grid over $[-10,10]\times[10,10]$; the apparent over-smoothed effect demonstrates the advantage of our bifidelity method in this setting. 
Right: Posterior distribution from an $41\times 41$ grid over the prior domain, which agrees well with the posterior obtained by our bifidelity method.
}
\label{fig:sde_comp}
\end{figure}

The quadratic nature of functions $h$ and $g$ implies that the target density should exhibit four equal modes centered at $(\pm\mu,\pm\sigma)$. We demonstrate our method's performance using two test cases with observations $\bm{y}_1 = [3.081 \ 0.658]^\top$ and $\bm{y}_2 = [1.462 \ 0.059]^\top$, corresponding to true parameters $\theta = [2\ 1]^\top$ and $\theta = [1.2 \ 0.3]^\top$, respectively.
Figs.~\ref{fig:SDE_1} and \ref{fig:SDE_2} demonstrate the accuracy of our approach compared to MCMC and SMC sampling; for this comparison, we employ a surrogate model in the likelihood computation, specifically a kernel regression estimator with a Gaussian kernel with a bandwidth selection of $10^{-4}$. This surrogate is trained on $201 \times 201$ samples from the parameter domain $\Gamma$, using $\Delta\mu = \Delta \sigma = 0.1$ spatial resolution and the high-fidelity simulation with $N_{\rm MC} = 10,000$. The figures clearly show that our score-based method successfully captures the expected four-fold symmetry of the modes, while MCMC and SMC struggle to identify this symmetric structure in the posterior distribution.
The computational efficiency of our approach is summarized in Table~\ref{tab::SDE_table}, which presents an example of wall-clock times for $\bm{y}_1$. The results demonstrate that our method achieves  computational savings for posterior sampling, while maintaining accuracy in sampling from the target posterior distribution. This point is further illustrated in Figure~\ref{fig:sde_comp}, demonstrating the computational advantage of first employing the low-fidelity solver and generative model, as compared to running the generative process directly with the high-fidelity simulation prior data in the baseline procedure.

\begin{table}[H] 
\centering
\renewcommand{\arraystretch}{1.25} 
\begin{tabular}{lccc}
\toprule
\small
\textbf{Model} & \textbf{Data labeling} & \textbf{Training $G^{(\cdot)}$} & \textbf{Synthesizing 10K samples} \\
\midrule
$G^{\rm low}$    & 0.65 seconds  & 15.55 seconds & 1.54e-03 seconds \\
\midrule
$G^{\rm high}|\bm{y}_1$   & 3.79 seconds & 14.54 seconds & 8.61e-04 seconds \\
$\text{MCMC}$   & N/A & N/A & 74.64 seconds \\
$\text{SMC}$   & N/A & N/A & 119.23 seconds \\
\bottomrule
\end{tabular}
\caption{Wall-clock run-times for different computational stages of the Ornstein-Uhlenbeck example. Data labeling refers to running the reverse-time ODE for training data 
For data labeling, we used $N_{\rm refine} = 1500$ samples for the observation $\bm{y}_1$. 
Training $G^{(\cdot)}$ refers to the training time of the neural network model (here the network architectures are taken to be 3 layers, 100 neurons each, and are trained for $10,000$ epochs). Synthesizing 10K samples refers to the run-time for neural network $G^{(\cdot)}$ (MCMC, SMC) to generate 10K samples. The results demonstrate the computational efficiency of our method in generating parameter samples for the stochastic differential equation system.}
\label{tab::SDE_table}
\end{table}
%
%

\subsection{Runaway electron (RE) dissipation with impurity injection}\label{sec:ex5}
Runaway electrons (REs) pose a significant threat to the safety and integrity of fusion reactors due to their high energies, which can cause severe damage to the reactor's walls and components \cite{del2021generation}.
One effective method to mitigate the dangers posed by REs is the use of impurity injection, e.g., shattered pellet injection (SPI) \cite{shiraki2018dissipation}. SPI introduces high-Z impurities into the plasma, rapidly increasing the electron density and encouraging RE dissipation through collisions. The Fokker-Planck (FP) differential with collision operators describes the evolution of electron distribution function $f(p,\eta,r,t)$ in momentum magnitude $p$, pitch angle $\eta$, and minor radius $r$,
\begin{equation}\label{eq:fp}
\mathcal{L}[f(p,\eta,r,t, \theta);\theta] = \frac{\partial f}{\partial t} - \mathcal{C}(f) + \mathcal{E}(f) + \mathcal{R}(f) + \mathcal{S}_{\rm Aval.}(f)=0,
\end{equation}
where $\mathcal{C}(f)$ is Coulomb (small-angle) collision operator, $\mathcal{E}(f)$ is the electric field acceleration, $\mathcal{R}(f)$ is synchrotron radiation damping, and $\mathcal{S}_{\rm Aval.}(f)$ is the large-angle collision operator for secondary RE generation. 

The numerical simulation of the Fokker-Planck equation in Eq.~\eqref{eq:fp} is performed using the Kinetic Orbit Runaway electrons Code (KORC) collision operator Monte-Carlo (MC) solver \cite{carbajal2017space}. We adopt a magnetic geometry with a major radius of $R_0 = 2.616$ (m), an on-axis toroidal magnetic field of $B_0 = 3.388$ (T), and a background electron temperature of $T_e = 2$ (eV); more details of the model can be found in the Appendix. To simplify the numerical simulation, we assume the impurity density for the $\alpha$-species is time-independent and spatially homogeneous. We focus on two impurity species: argon (Ar) and deuterium (D), considering both their neutral forms ${\rm Ar}$ and ${\rm D}$ and their partially ionized charge states ${\rm Ar}^{+1}$ and ${\rm D}^{+1}$. For the neutral impurities, we assume constant fixed values of $n_{{\rm Ar}} = n_{{\rm D}} = 1 \times 10^{19}  (\text{m}^{-3}$), while treating $n_{{\rm Ar}^{+1}}$ and $n_{{\rm D}^{+1}}$ as input variables for the study. Additionally, we assume a spatially and temporally homogeneous electric field $E$ serving as another input variable. In this case, equal impurity densities are applied to both ${\rm Ar}^{+}$ and ${\rm D}^{+}$ ions, with ${n}_{{\rm Ar}^{+}}={n}_{{\rm D}^{+}} = n_{\rm imp}$. Cases with different impurity densities will be investigated in future work.

Consequently, two input parameters $\{\theta^{(i)}\}_{i=1}^{N}=\{ n^{(i)}_{\rm imp}, E^{(i)}\}_{i=1}^{N}$ are varied for the study.
For a given simulation input parameter set $\theta^{(i)}$, we evolve RE populations $\{(p^{(n)},\xi^{(n)},r^{(n)})\}_{n=1}^{N_{\rm MC}}$ by running KORC Monte-Carlo solver. The initial RE condition corresponds to a 5 MeV monoenergetic, $10^\circ$ monopitch and the runaway electron density is spatially constant in the normalized radius $r$. We step the simulation time as $T=20$ ms. 
The avalanche term $\mathcal{S}_{\rm Aval.}(f)$ significantly increases the computational cost of sampling-based numerical methods due to the exponential growth of secondary REs. Therefore, it is essential to explore the bifidelity diffusion model to efficiently capture the detailed dynamics in the region of interest.  

To characterize the RE ensemble evolution at a macroscopic level, we define the total RE current $I_{\rm RE}$ as \cite{beidler2020spatially} 
\begin{equation}
\begin{aligned}
& I_{\rm RE} = -\frac{e}{2\pi} \sum_n^{N_{\rm MC}} \frac{\nu_{\phi}}{R_n} \mathcal{M}_{\rm RE,n}, \quad \text{where} \, \mathcal{M}_{\rm RE,n} = \begin{cases}
 1, \quad &\text{if}\quad p_n > m_e c \\
 0, \quad &\text{if}\quad p_n < m_e c, \quad  \text{or}\quad  \text{hits the wall},
 \end{cases}
\end{aligned}
\end{equation}
where $\nu_{\phi} \simeq \nu \cos\eta$ with pitch angle $\eta$.
The output of interest  is the normalized  RE current  damping rate $\varphi_{I_{\rm RE}} = \frac{d I_{\rm RE}}{dt} \approx\frac{I_{\rm RE}(T) - I_{\rm RE}^0}{T}$, which is numerically calculated by linear fitting\footnote{In our work, we normalize the RE current $I_{\rm RE}$ by its initial value $I_{\rm RE}^0$, i.e., $\bar{I}_{\rm RE} = {{I}_{\rm RE}}/{{I}_{\rm RE}^0}$. For the simplicity of notation, we omit the overbar for the normalized quantities.}. Overall, for location $X = (p,\eta, r)$, the observational operator is $\mathcal{H}({X}(\theta))$,
where the output is the damping rate $\varphi_{I_{\rm RE}}$, and input parameters are ${\theta} = (n_{\rm imp}, E)$.
Let $N_{\rm MC}$ denote the number of trajectories of Monte Carlo solver where $N_{\rm MC}$ is an adjustable hyper-parameter of the simulation model. It follows that the Monte Carlo approximation of the observational operator can be written as

\begin{equation} \label{eq:RE_MC_error_eqn}
    y = \mathcal{H}^{\rm MC}(X(\theta)) = \mathcal{H}(X(\theta)) + \epsilon,
\end{equation}
where $y$ the noisy observation of $\varphi_{\rm I_{RE}}$ and $\mathbf{\epsilon} \sim \mathcal{N}(\mathbf{0},\Sigma)$ for some covariance matrix $\Sigma$. We assume $\mathcal{H}^{\rm MC}(X(\theta))$ is an asymptotically unbiased estimator of $\mathcal{H}(X(\theta))$ i.e. $\mathbb{E}[\mathcal{H}^{\rm MC}(X(\theta))] = \mathcal{H}(X(\theta))$ as $N_{\rm MC} \rightarrow \infty$. Our goal is to compute the conditional distribution $p((n_{{\rm imp}}, E)|\varphi_{I_{\rm RE}})$ given a target damping rate $\varphi_{I_{\rm RE}}$. 

\begin{figure}[h!]
    \centering
\includegraphics[width=0.5\textwidth]{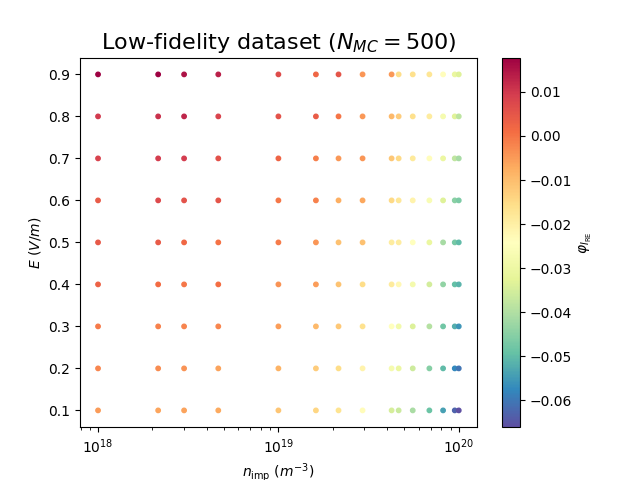}
\includegraphics[width=\textwidth]{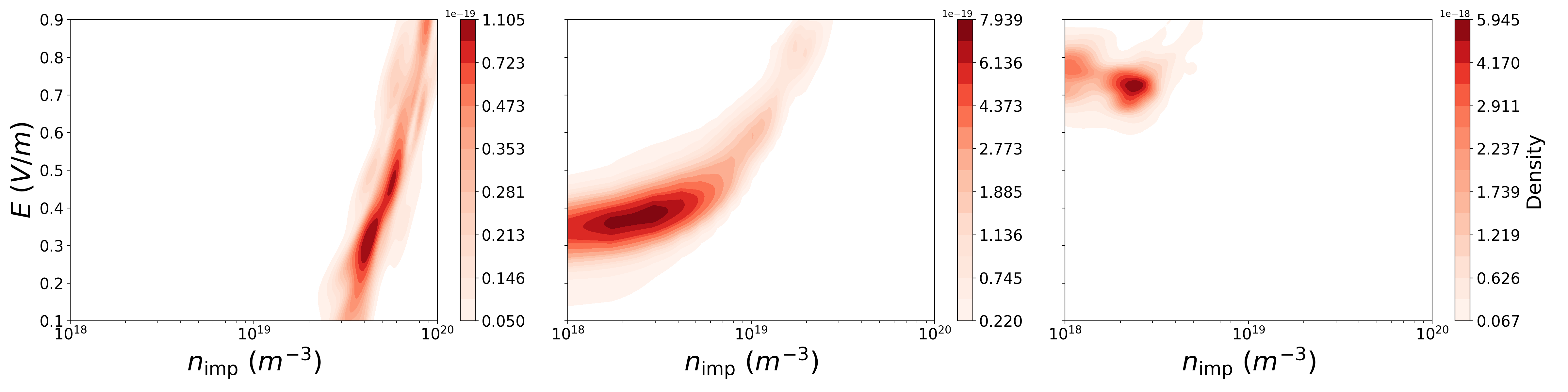}
\caption{$\varphi_{I_{\rm RE}}$ with varying $n_{{\rm imp}}$ and $E$ over parameter domain $\Gamma$. The training data set is consist of 135 simulation (generated 15 x 9 grid) shots, with a logarithmic scale in $n_{\rm imp}$ and a linear scale in $E$. Each simulation shot is performed using $N_{\rm MC}=500$ electron trajectories.}
 \label{fig_EC_Nimp}
\end{figure}

For training the low-fidelity model $G^{\rm low}$, we first generate data samples using $N_{\rm MC} = 500$ simulation paths to obtain low-fidelity outputs. We take our initial sampling domain for $(n_{\rm imp},E)$ to be a $15 \times 9$ uniform grid over the domain $\Gamma = [1e19, 1e20] \times [0.1, 0.9]$ depicted in Fig.~\ref{fig_EC_Nimp}. Hence our low-fidelity prior data set is $\mathcal{S}_{\rm prior} = \{({\theta}^{(i)},{y}^{(i)})\}_{i=1}^{N_{\rm prior}}$ with $N_{\rm prior} = 135$. It illustrates the effects of the impurity density $n_{\rm imp}$ ($x$-axis) and electric field magnitude $E$ ($y$-axis) on the RE current decay rate $\varphi_{I_{\rm RE}}$. As the electric field amplitude increases, more runaway electrons are generated due to stronger acceleration, resulting in a higher RE current. Conversely, as the impurity density increases, collisions between runaway electrons and impurity ions become more frequent, leading to enhanced energy dissipation and a reduction in RE current.
To compute the likelihood in the score function approximation, we require an estimate of $\Sigma$ in Eq.~\eqref{eq:RE_MC_error_eqn}, which we approximate using Eq.~\eqref{variance_estimator_SDE_example} with $N_{\rm grid} = 20$. Since we do not have access to the ground truth damping rate, we approximate reference values using $N_{\rm MC} = 60000$. We take the architecture of all generative models in this example to be a fully connected feed-forward network with two hidden layers, 100 neurons each. 


\begin{figure}[h!] 
    \centering
\includegraphics[width=\linewidth]{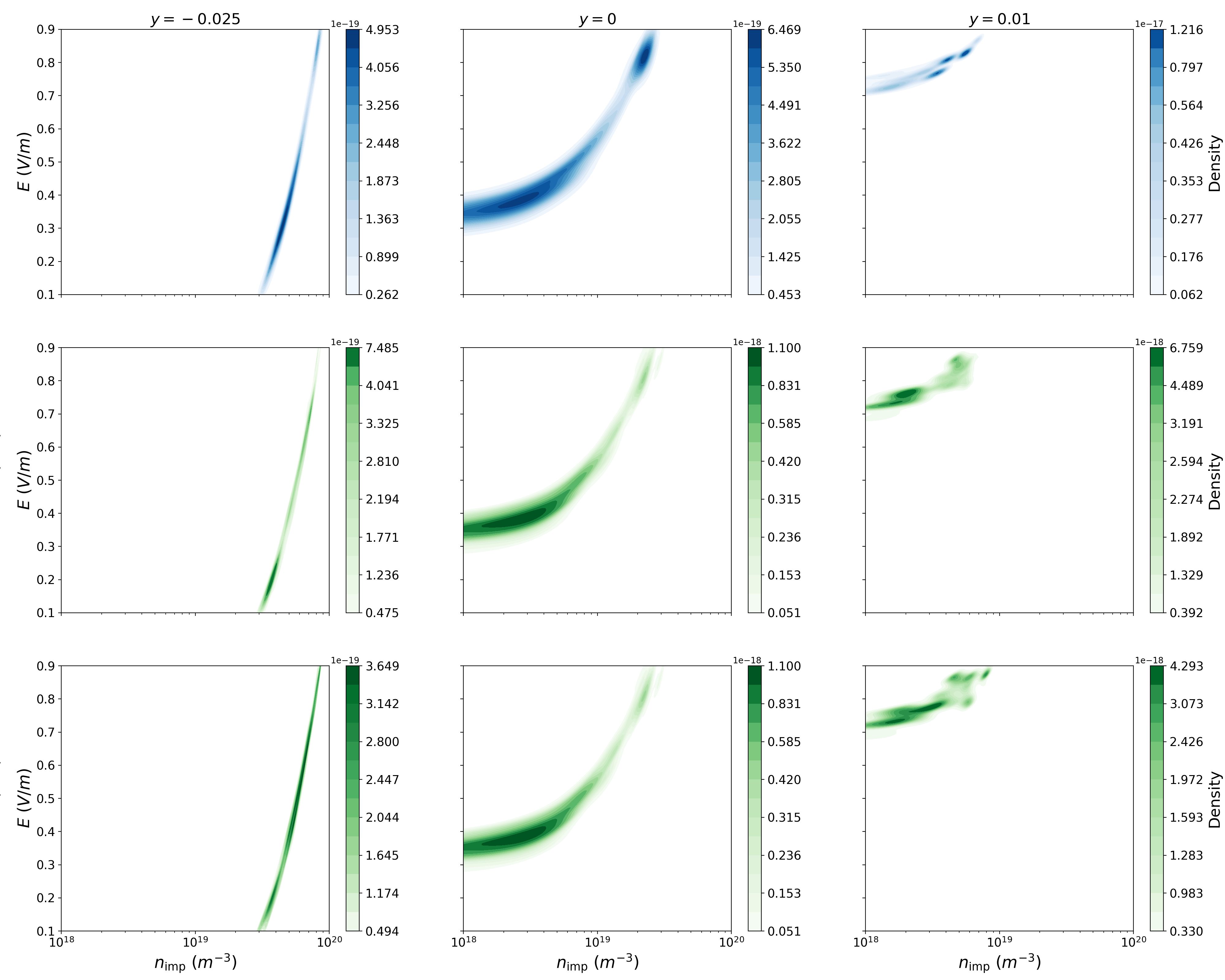}

\caption{{Run-away electron example.}The high-fidelity solver explores the target domain for $\varphi^*_{I_{\rm RE}}= 0.01, 0, -0.025$ using more simulation shots with $N_{\rm MC}=5000$ electron trajectories per case. Top row: The data samples in $\mathcal{S}_{\rm refine}$ generated from the range of $G^{\rm low}$ samples. Second row: The contourf plot shows the distribution $p((n_{{\rm imp}}, E)|\varphi_{I_{\rm RE}})$ for target RE current damping rate with $\varphi^*_{I_{\rm RE}}=0.01, 0, -0.025$ (from left to right), generated from the labeled data sets. 
The second iteration results in a refined density approximation of the target distribution. Third row: The contourf plots generated by $G^{\rm high}$, allowing for efficient data sampling from the target distribution. Bottow row: The corresponding results from the MCMC algorithm using a surrogate model and 10 walkers within the refined domain (restricted to range of samples generated by ${G}^{\rm low}$).
We see good agreement between the results of the diffusion model and MCMC.}
\label{fig:RE_2}
\end{figure}

Fig.~\ref{fig:RE_2} illustrates the distribution
$p((n_{{\rm imp}, }, E)|\varphi_{I_{\rm RE}})$ for the target RE current damping rate with specific values of $\varphi^*_{I_{\rm RE}}=0.01, 0, -0.025$ (from left to right). Achieving higher accuracy in specific regions requires further refinement using the high-fidelity solver.


{We then employ the high-fidelity solver to examine the target domain in greater detail. The high-fidelity model conducts additional simulations within this domain, using $N_{\rm MC}=5000$ electron trajectories per simulation to construct $\mathcal{S}_{\rm refine}$. To determine the number of $N_{\rm refine}$ samples, we employ a refinement schedule beginning at 100 samples and increasing by increments of 50. We again select our tolerance as $\varepsilon^{\rm TOL} = 10^{-2}$. For all observations, this criterion is met at $N_{\rm refine} = 200$. }

\begin{figure}
    \centering
    \includegraphics[width=\linewidth]{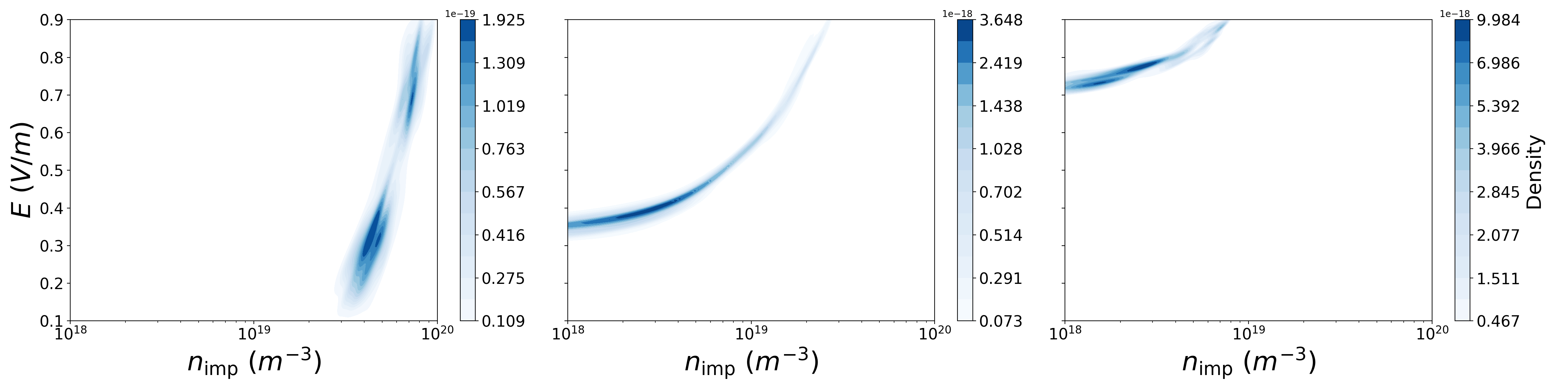}
    \caption{{Run-away electron example.} Posterior distributions generated by running the baseline procedure on the $31 \times 17$ finer resolution grid. We note good agreement between the posterior distributions conditioned on $\varphi^*_{I_{\rm RE}}=0, 0.01$ with the bifidelity and MCMC/SMC results, however an over-smoothed result of $\varphi^*_{I_{\rm RE}}= - 0.025$. This suggests that the finer resolution allows for good approximations for some conditions, but may require more samples for other observations of interest;  this highlights an important benefit of our two-stage method, which offers computational savings in the first step, as $G^{\rm low}$ can be used for posterior approximations over many observations, while allowing for the adaptive refinement for observations of interest in the second step.}
    \label{fig:RE_HF_comp}
\end{figure}

Figure~\ref{fig:RE_2} shows the results for $\varphi^*_{I_{\rm RE}}=-0.025, 0, 0.01$. The top row depicts results from the high-fidelity generative model $G^{\rm high}$. The middle and bottom row demonstrates a comparison with MCMC and SMC respectively, where for the likelihood evaluations, a kernel regression estimator is employed, constructed from high-fidelity data obtained at a finer resolution grid over the prior domain, specifically a $31 \times 17$ grid for a total of 527 samples. We observe good agreement between all results. In Figure~\ref{fig:RE_HF_comp}, using the same fine resolution grid, we further demonstrate the posterior density approximations generated using our baseline procedure for each of the observations of interest. We observe good agreement between the bifidelity and the baseline procedure results. We note that while the bifidelity model requires $N_{\rm refine} = 200$ runs for each of the observations, the high-fidelity baseline, MCMC and SMC required 527 forward solves. Table~\ref{tab::RE_table} presents the computational run-times for each stage of our method. The results demonstrate computational efficiency in generating posterior samples compared to MCMC and SMC.


\renewcommand{\arraystretch}{1.25}
\begin{table}[h!] 
\centering
\small
\begin{tabular}{lccc}
\toprule
\textbf{Model} & \textbf{Data labeling} & \textbf{Training $G^{(\cdot)}$} & \textbf{Synthesizing 10K samples} \\
\midrule
$G^{\rm low}$    & 1.76  seconds  & 8.12 seconds &  4.30e-04 seconds \\
\midrule
$G^{\rm high}, y = 0$   & 0.93 seconds & 8.77 seconds & 4.50e-04 seconds \\
$\text{MCMC}$   & N/A & N/A &  16.54 seconds
\\
$\text{SMC}$   & N/A & N/A &  28.66 seconds \\
\bottomrule
\end{tabular}
\caption{Wall-clock run-times for different computational stages of the runaway electron example. Data labeling refers to running the reverse-time ODE for training data 
 Training $G^{(\cdot)}$ refers to the training time of the neural network model (3 hidden layers, 100 neurons per layer, trained for $10,000$ epochs). Synthesizing 10K samples refers to the run-time for neural network $G^{(\cdot)}$ (MCMC) to generate 10K samples. For $G^{\rm low}$, this is the average run time to synthesize the 10K samples for the three observations of interest. The results demonstrate the computational efficiency of our method in generating parameter samples for the runaway electron system.} 
\label{tab::RE_table}
\end{table}

\section{Conclusion}\label{sec:conclusion}
We introduced a bifidelity method for obtaining posterior distributions of parameter estimates conditioned on data observations of interest. This method leverages low- and high-fidelity models in conjunction with a training-free score-based diffusion model to efficiently generate the posterior density approximation. Supervised learning of neural networks is used to develop low- and high-fidelity generative models capable of efficiently producing samples from the posterior distribution. The conditional low-fidelity generative model enables amortized Bayesian inference for efficient density approximations over a wide range of data observations. The high-fidelity generative model activates selectively when higher accuracy density approximations are required for a specific data observation, utilizing the outputs from the low-fidelity generative model to efficiently sample from the parameter space for building of the high-fidelity training data set. Once trained, the high-fidelity generative model is capable of accurately approximating the target posterior distribution. The effectiveness of the proposed method is demonstrated through several examples of complex systems, including parameter estimation for a plasma physics application aimed at mitigating runaway electrons, critical for ensuring the safety of fusion reactors. When analytical reference distributions are unavailable, we compare results and wall-clock run-times with MCMC and sequential Monte Carlo, two commonly used algorithms for posterior density sampling, to highlight the accuracy and efficiency of our method. This flexible framework also provides a foundation for extensions to multifidelity applications in parameter estimation, inverse problems, and forward state estimation problems for higher-dimensional physical systems, which will be explored in future studies.

\section*{Acknowledgment}
This work is supported by the U.S. Department of Energy, Office of Science, Office of Advanced Scientific Computing Research, Applied Mathematics program, under the contract ERKJ443, and accomplished at Oak Ridge National Laboratory (ORNL), and under Grant DE-SC0022254. This material is also partially supported by the U.S. Department of Energy, Office of Science, Office of Workforce Development for Teachers and Scientists, Office of Science Graduate Student Research (SCGSR) program. The SCGSR program is administered by the Oak Ridge Institute for Science and Education for the DOE under contract number DE-SC0014664. ORNL is operated by UT-Battelle, LLC., for the U.S. Department of Energy under Contract DE-AC05-00OR22725.

\section*{Appendix}
\addcontentsline{toc}{section}{Appendix} 
The Coulomb collisions are a crucial mechanism for runaway electron (RE) dissipation. These collisions, between REs and injected impurities, play a key role in dissipating the energy of REs by increasing their interaction rates with the surrounding plasma. The Coulomb collision operator is defined as 
\begin{align*}
\mathcal{C}(f) = \frac{1}{p^2}\frac{\partial}{\partial p}\left[p^2 \left(C_A\frac{\partial f}{\partial p} + C_F f\right) \right] + \frac{\mathcal{C}_B}{p^2}\left[\frac{1}{\sin{\eta}} \frac{\partial}{\partial \eta}\left(\sin{\eta} \frac{\partial f}{\partial \eta} \right) \right],
\end{align*}
where $C_F$, $C_A$ and $C_B$ are linearized transport coefficients for collisional friction, parallel diffusion, and pitch angle scattering, separately. 
We integrate collision operators by generalizing the coefficients of these operators to include both non-relativistic and relativistic energy limits. With the model of bound electrons and partially ionized impurities \cite{hesslow2018generalized}, the transport coefficients are 
\begin{align*}
\begin{aligned}
C_{A, {\rm Bound}}(\nu) =& \frac{\Gamma_{ee} \mathcal{G}\left(\frac{\nu}{\nu_{\rm th}}\right)}{\nu}\\
C_{F, {\rm Bound}}(\nu) =& \frac{\Gamma_{ee} \mathcal{G}\left(\frac{\nu}{\nu_{\rm th}}\right)}{T_e} \left\{ 1+\sum_{j} \frac{n_j}{n_e}\frac{Z_j - Z_{0j}}{\ln{ \Lambda_{ee}}} \left[\frac{1}{5}\ln{(1+h_j^5)} - \beta^2 \right] \right\}\\
C_{B, {\rm Bound}}(\nu) =& \frac{\Gamma_{ei}}{2\nu} Z_{\rm eff}\left(1+\frac{1}{Z_{\rm eff}} \sum_j \frac{n_j}{n_e}\frac{g_j}{\ln{\Lambda_{ei}}}\right)\\
&+  \frac{\Gamma_{ee}}{2\nu}\left[{\rm erf}\left(\frac{\nu}{\nu_{\rm th}}\right) - \mathcal{G}\left(\frac{\nu}{\nu_{\rm th}}\right) + \frac{1}{2} \left(\frac{\nu_{\rm th} \nu}{c^2} \right)\right],
\end{aligned}
\end{align*}
where $\Gamma_{ee,ei} = n_e e^4 \ln \Lambda_{ee,ei} /4\pi \epsilon_0^2$ with
$\ln \Lambda_{ee,ei}$ is the Coulomb logarithm for e-e (e-i) collisions, $\nu_{\rm th} = \sqrt{2T_e/m_e}$ is the thermal electron velocity, and $\mathcal{G}$ is the Chandrasekhar function.
$n_j$ is the density of the $j$-th ionization state, $Z_j$ is the unscreened (i.e., fully ionized) impurity ion charge, and $Z_{0,j}$ is the screened (i.e., partially ionized) impurity ion charge. $h_j = p\sqrt{\gamma - 1}/(m_e c I_j)$, where $I_j$ is the mean excitation energy and $\beta = \nu/c$ is the relativistic speed. The term $Z_j - Z_{0j}$ is used to define the number of bound electrons a partially ionized impurity charge state possesses. Further details on this model can be found in Refs.~\cite{hesslow2017effect,beidler2024wall,boozer2015theory,embreus2018relativistic,mcdevitt2019avalanche} and references therein.

\bibliographystyle{siamplain}
\bibliography{combined_references}

\end{document}